\documentclass[10pt,journal,cspaper,compsoc]{IEEEtran}
\usepackage{geometry}
\usepackage{amssymb,amsmath}
\usepackage{color}
\usepackage{multirow}
\usepackage{booktabs}
\usepackage{mathrsfs}
\usepackage{bm}
\newcommand{\tabincell}[2]{\begin{tabular}{@{}#1@{}}#2\end{tabular}}

\ifCLASSOPTIONcompsoc
\else
\fi
\usepackage{graphicx}
\usepackage{subfigure}
\usepackage{rotating}
\usepackage{threeparttable}
\ifCLASSINFOpdf
\else
\fi
\begin{document}
%
\title{Trunk-Branch Ensemble Convolutional Neural Networks for Video-based Face Recognition}
%
%
%
%
\author{Changxing~Ding,~\IEEEmembership{Member,~IEEE},
        Dacheng~Tao,~\IEEEmembership{Fellow,~IEEE}
\thanks{C. Ding is with the School of Electronic and Information Engineering,
South China University of Technology, 381 Wushan Road, Tianhe District, Guangzhou 510000, P.R.China.
(Email: chxding@scut.edu.cn).}
\thanks{D. Tao is with the UBTech Sydney Artificial Intelligence Institute and the School of Information Technologies
in the Faculty of Engineering and Information Technologies at The University of Sydney, Darlington, NSW 2008, Australia.
(Email: dacheng.tao@sydney.edu.au).}}



%
%

\markboth{IEEE TRANSACTIONS ON PATTERN ANALYSIS AND MACHINE INTELLIGENCE,~Vol.X, No.X, MONTH~2017}%
{DING \MakeLowercase{\textit{et al.}}: Trunk-Branch Ensemble Convolutional Neural Networks for Video-based Face Recognition}


%


\IEEEcompsoctitleabstractindextext{%
\begin{abstract}
Human faces in surveillance videos often suffer from severe image blur, dramatic pose variations, and occlusion.
In this paper, we propose a comprehensive framework based on Convolutional Neural Networks (CNN) to overcome challenges in video-based face recognition (VFR).
First, to learn blur-robust face representations,
we artificially blur training data composed of clear still images to account for a shortfall in real-world video training data.
Using training data composed of both still images and artificially blurred data, CNN is encouraged to learn blur-insensitive features automatically.
Second, to enhance robustness of CNN features to pose variations and occlusion, we propose a Trunk-Branch Ensemble CNN model (TBE-CNN),
which extracts complementary information from holistic face images and patches cropped around facial components.
TBE-CNN is an end-to-end model that extracts features efficiently by sharing the low- and middle-level convolutional layers between the trunk and branch networks.
Third, to further promote the discriminative power of the representations learnt by TBE-CNN,
we propose an improved triplet loss function.
Systematic experiments justify the effectiveness of the proposed techniques.
Most impressively, TBE-CNN achieves state-of-the-art performance on three popular video face databases:
PaSC, COX Face, and YouTube Faces.
With the proposed techniques, we also obtain the first place in the BTAS 2016 Video Person Recognition Evaluation.
\end{abstract}

\begin{keywords}
Video-based face recognition, video surveillance, blur- and pose-robust representations, convolutional neural networks.
\end{keywords}}

\maketitle


%

\section{Introduction}
\label{Sec:Introduction}
%
%

%
%
%
%

\IEEEPARstart{W}{ith} the widespread use of video cameras for surveillance and mobile devices, an enormous quantity of video is constantly being captured.
Compared to still face images, videos usually contain more information, e.g., temporal and multi-view information.
The ubiquity of videos offers society far-reaching benefits in terms of security and law enforcement.
It is highly desirable to build surveillance systems coupled with face recognition techniques to automatically identify subjects of interest.
Unfortunately, the majority of existing face recognition literature focuses on matching in still images,
and video-based face recognition (VFR) research is still in its infancy~\cite{Huang2015Benchmark,best2013video}.
In this paper, we handle the still-to-video (S2V), video-to-still (V2S), and video-to-video (V2V) matching problems,
which are used in the most common VFR applications.

\begin{figure}
\centering
\includegraphics[width=1.0\linewidth]{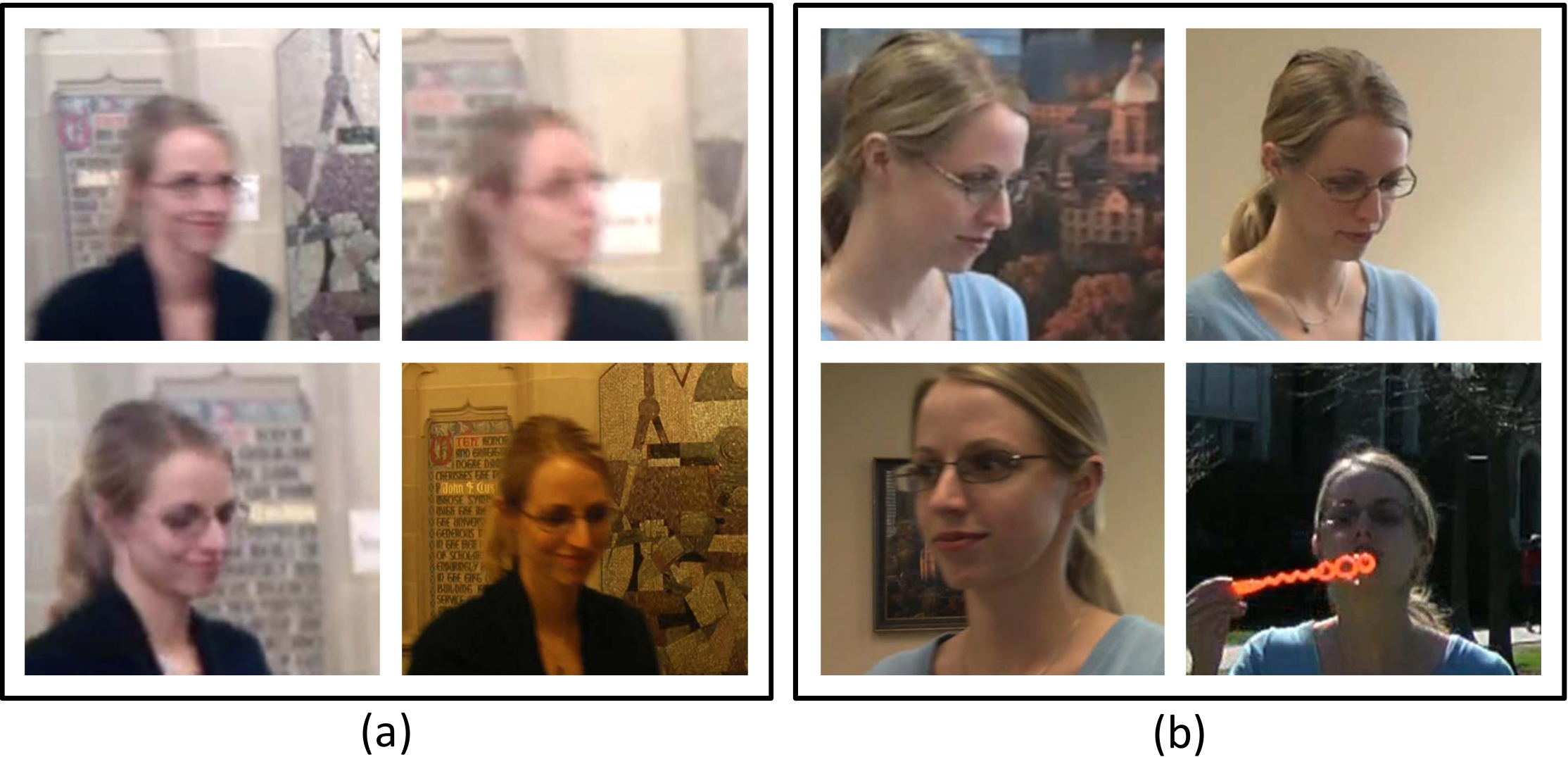}
\caption{Video frames captured by surveillance or mobile devices suffer from severe image blur, dramatic pose variations, and occlusion.
(a) Image blur caused by the motion of the subject, camera shake (for mobile devices), and out-of-focus capture.
(b) Faces in videos usually exhibit occlusion and a large range of pose variations.
}
\label{fig:BlurPose}
\end{figure}

Compared to still image-based face recognition (SIFR), VFR is significantly more challenging.
Images in standard SIFR datasets are usually captured under good conditions or even framed by professional photographers, e.g., in the Labeled Faces in the Wild (LFW) database~\cite{LFWTech}.
In comparison, the image quality of video frames tends to be significantly lower and faces exhibit much richer variations (Fig.~\ref{fig:BlurPose})
because video acquisition is much less constrained.
In particular, subjects in videos are usually mobile, resulting in serious motion blur, out-of-focus blur, and a large range of pose variations.
Furthermore, surveillance and mobile cameras are often low-cost (and therefore low-quality) devices, which further exacerbates problems with video frame clarity~\cite{biswas2013pose}.

Recent advances in face recognition have tended to ignore the peculiarities of videos when extending techniques from SIFR to VFR~\cite{schroff2015facenet,sun2015deeply,parkhi2015deep,li2015hierarchical}.
On the one hand, a major difficulty in VFR, such as severe image blur, is largely unsolved~\cite{Ross2015report}.
One important reason is that large amounts of real-world video training data are still lacking,
and existing still image databases are usually blur-free.
On the other hand, although pose variations and occlusion are partially solved in SIFR by ensemble modelling~\cite{sun2015deeply,liu2015targeting},
the strategy may not be directly extended to VFR.
A common practice in model ensembles is to train models separately for the holistic face image and for patches cropped around facial components.
Model fusion is then performed offline at the feature~\cite{sun2014deep} or score level~\cite{ding2014multi}.
However, the accuracy promoted by model ensembles is at the cost of significantly increased time cost,
which is impractical for VFR since each video usually contains dozens or even thousands of frames.

Here we approach the blur-robust representation learning problem from the perspective of training data.
Since the volume of real-world video training data is small,
we propose simulating large amounts of video frames from existing still face image databases.
During training, we provide CNN with two training data streams:
one composed of still face images, and the other composed of simulated video frames created by applying random artificial blur to the first stream.
The network aims to classify each still image and its artificially blurred version into the same class; therefore, the learnt face representations must be blur-insensitive.
To the best of our knowledge, this is the first CNN-based approach to solve the image blur problem in VFR.

To learn pose- and occlusion-robust representations for VFR efficiently,
we propose a novel end-to-end ensemble CNN model called Trunk-Branch Ensemble CNN (TBE-CNN).
TBE-CNN includes one trunk network and several branch networks. The trunk network learns face representations for holistic face images,
and each branch network learns representations for image patches cropped around one facial component.
To speed up computation, the trunk and branch networks share the same low- and middle-level layers,
while their high-level layers are optimized independently.
This sharing strategy significantly reduces the computational cost of the ensemble model and at the same time exploits each model's uniqueness.
The output feature maps by the trunk network and branch networks are fused by concatenation to form a comprehensive face representation.

Furthermore, to enhance TBE-CNN's discriminative power, we propose a Mean Distance Regularized Triplet Loss (MDR-TL) function to train TBE-CNN in an end-to-end fashion.
Compared with the popular triplet loss function~\cite{schroff2015facenet}, MDR-TL takes full advantage of label information
and regularizes the triplet loss so that inter- and intra-class distance distributions become uniform.

The efficacy of the proposed algorithm is systematically evaluated on three popular video face databases:
PaSC~\cite{beveridge2013challenge}, COX Face~\cite{Huang2015Benchmark}, and YouTube Faces~\cite{wolf2011face}.
The evaluation is conducted for S2V, V2S, and V2V tasks.
Extensive experiments on the three datasets indicate that the proposed algorithm achieves superior performance.

The remainder of the paper is organized as follows. Section~\ref{Sec:Related} briefly reviews related VFR works.
The TBE-CNN model that handles image blur and pose variations for VFR is described in Section~\ref{Sec:TBECNN}.
The TBE-CNN training strategy is introduced in Section~\ref{Sec:Training}.
Face matching using the proposed approach is illustrated in Section~\ref{Sec:VFR}.
Results of experiments are presented in Section~\ref{Sec:EXPERIMENTS}, leading to conclusions in Section~\ref{Sec:Conclusion}.

\section{Related Works}
\label{Sec:Related}
We review the literature in two parts: 1) video-based face recognition, and 2) deep learning methods for face recognition.

\subsection{Video-based Face Recognition}
Existing studies on VFR can be divided into three categories:
(i) approaches for frame quality evaluation, (ii) approaches that exploit redundant information contained between video frames,
and (iii) approaches that attain robust feature extraction from each frame.

Frame quality evaluation is mainly utilized for key frame selection from video clips,
such that only a subset of best-quality frames is selected for efficient face recognition.
For a systematic summary of frame quality evaluation methods, we direct readers to two recent works~\cite{phillips2013existence,mau2013video}.

In contrast to frame quality evaluation methods, a number of recent VFR studies attempt to make use of redundant information contained between video frames.
Algorithms that fall in this category include sequence-based methods, dictionary-based methods, and image set-based methods~\cite{barr2012face}.
The sequence-based methods aim to extract person-specific facial dynamics from continuous video frames~\cite{bicego2006person,hadid2009combining},
which means that they rely on robust face trackers.
The dictionary-based methods construct redundant dictionaries using video frames
and employ sparse representation-based classifiers for classification~\cite{chen2012dictionary,liu2014toward}.
Due to the large size of the constructed dictionaries, the dictionary-based methods are often inefficient.
The image set-based methods model the distribution of video frames using various techniques, e.g., affine/convex hull~\cite{hu2012face,zhu2014image},
linear subspace~\cite{Huang2015Benchmark}, and manifold methods~\cite{harandi2011graph,cui2012image}.
They then measure the between-distribution similarity to match two image sets.
The downside of image set modeling is that it is sensitive to the variable volume of video frames
and complex facial variations that exist in real-world scenarios~\cite{shao2015comparative}.

Extracting high-quality face representations has always been a core task in face recognition~\cite{ding2015comprehensive}.
In contrast to still face images, video frames usually suffer from severe image blur because of the relative motion between the subjects and the cameras.
Two types of methods have been introduced to reduce the impact of image blur:
deblur-based methods~\cite{nishiyama2011facial} and blur-robust feature extraction-based methods~\cite{gopalan2012blur,chan2013multiscale}.
The former method first estimates a blur kernel from the blurred image and then deblurs the face image prior to feature extraction.
However, the estimation of the blur kernel is challenging, as it is an ill-posed problem.
Of the blur-robust feature extraction methods,
Ahonen \textit{et al.}~\cite{ahonen2008recognition} proposed to employ the blur-insensitive Local Phase Quantization (LPQ) descriptor for facial feature extraction,
which has been widely used in VFR applications~\cite{Ross2015report}.
However, to the best of our knowledge, no CNN-based method has yet been used to handle the image blur problem in VFR.
Furthermore, similar to still face images, faces in video frames exhibit rich pose, illumination, and expression variations and occlusion;
therefore, existing studies tend to directly extend feature extractors designed for SIFR to VFR~\cite{parkhi2014compact,li2014eigen,chen2016efficient}.

Our proposed approach falls into the third category of methods.
Compared to previous VFR approaches, we propose an efficient CNN model to automatically learn face representations
that are robust to image blur, pose variations, and occlusion.

\subsection{Deep Learning Methods for Face Recognition}
Representation learning with CNNs provides an effective tool for face recognition.
Promising results have been obtained, but they are usually limited to SIFR~\cite{schroff2015facenet,parkhi2015deep,roychowdhury2015face}.
For example, Taigman \textit{et al.}~\cite{taigman2014deepface} formulated face representation learning as a face identification problem in CNN.~\cite{sun2014deep,hu2014discriminative,schroff2015facenet,lu2015multi} proposed deep metric learning methods to enhance the discriminative power of learnt face representations.
In comparison, there are only limited studies on CNN-based VFR.

This is for a number of reasons. First, existing CNN-based approaches do not handle the peculiarities of video frames (such as severe image blur) very well.
Training from real-world video data is an intuitive solution for VFR.
However, existing face video databases are rather small, e.g., 2,742 videos in PaSC and 3,000 videos in COX Face.
Second, the video data are highly redundant, because frames in the same video tend to be similar in image quality, expression, occlusion, and illumination conditions.
Therefore, direct CNN training using real-world video data is prone to overfitting.
As a result, the majority of existing works employ CNN models trained on large-scale still image databases for VFR~\cite{taigman2014deepface,parkhi2015deep}
and ignore the difference in image quality between still images and video frames.
Recently, Huang \textit{et al.}~\cite{Ross2015report} proposed pre-training CNN models with a large volume of still face images
and then fine-tuning the CNN models with small real-world video databases.
However, the fine-tuning strategy is suboptimal, as it only slightly adapts CNN parameters to the video data.
In comparison, our proposed video data simulation strategy enables direct optimization of all model parameters for VFR.
Video data simulation can be regarded as a novel method for data augmentation.
Compared to previous augmentation methods such as horizontal flipping and translation,
video data simulation produces visually different images compared to the original still images, as illustrated in Fig.~\ref{fig:BlurredImages}.

Part-based models have been widely used in both detection~\cite{ouyang2013joint,savalle2014deformable,girshick2015deformable,wan2015end,yang2016end}
and recognition~\cite{zhang2014part,zhang2015fine,huang2016part,lin2015bilinear,max2015spatial} tasks to account for occlusion, pose variations, and non-rigid object deformations.
For SIFR~\cite{sun2014deep,liu2015targeting},
the principle is to train CNN models separately using the holistic face image and patches cropped around facial components.
Model fusion is usually conducted offline by directly concatenating representations learnt by all models.
Since image patches are less sensitive to pose variations and occlusion~\cite{arashloo2011energy}, the ensemble system outperforms single models.
However, there are drawbacks from two perspectives. First, the ensemble system is not an end-to-end system,
which means its parameters are not simultaneously optimized for the VFR task.
Second, extracting features separately from multiple models significantly reduces efficiency,
which may be impractical for VFR since each video tends to have a number of frames to process.

A few recent works for fine-grained image classification are relevant to this paper~\cite{zhang2014part,zhang2015fine,huang2016part}.
Compared with~\cite{huang2016part}, TBE-CNN is more concise and more efficient:
it integrates the networks for the holistic image and all facial components into a single model by sharing their low- and middle-level layers.
Besides,~\cite{zhang2014part,zhang2015fine,huang2016part} dictate that object parts share all convolutional layers
and a representation of each part is formed by cropping feature maps from the last convolutional layer of CNN.
However, feature maps extracted by the high-level convolutional layers represent abstract and global information;
therefore, it may be less effective to utilize it to compute the representations of facial components.
In comparison, TBE-CNN is more discriminative: the holistic image and each facial component have independent high-level convolutional layers to better exploit their complementary information.

\section{Trunk-Branch Ensemble CNN}
\label{Sec:TBECNN}
In this section, we make two contributions to VFR.
First, we introduce a method for artificially simulating video-like training image data for blur-robust face recognition.
Second, the efficient TBE-CNN model for VFR is described.
As we lack real-world video sequences, the training set of TBE-CNN is composed of a large amount of face images.
For testing, TBE-CNN is employed to extract features from individual video frames and fuse their features by average pooling,
detailed in Section~\ref{Sec:VFR}.

\subsection{Artificially Simulated Video Data}
\begin{figure}
\centering
\includegraphics[width=1.0\linewidth]{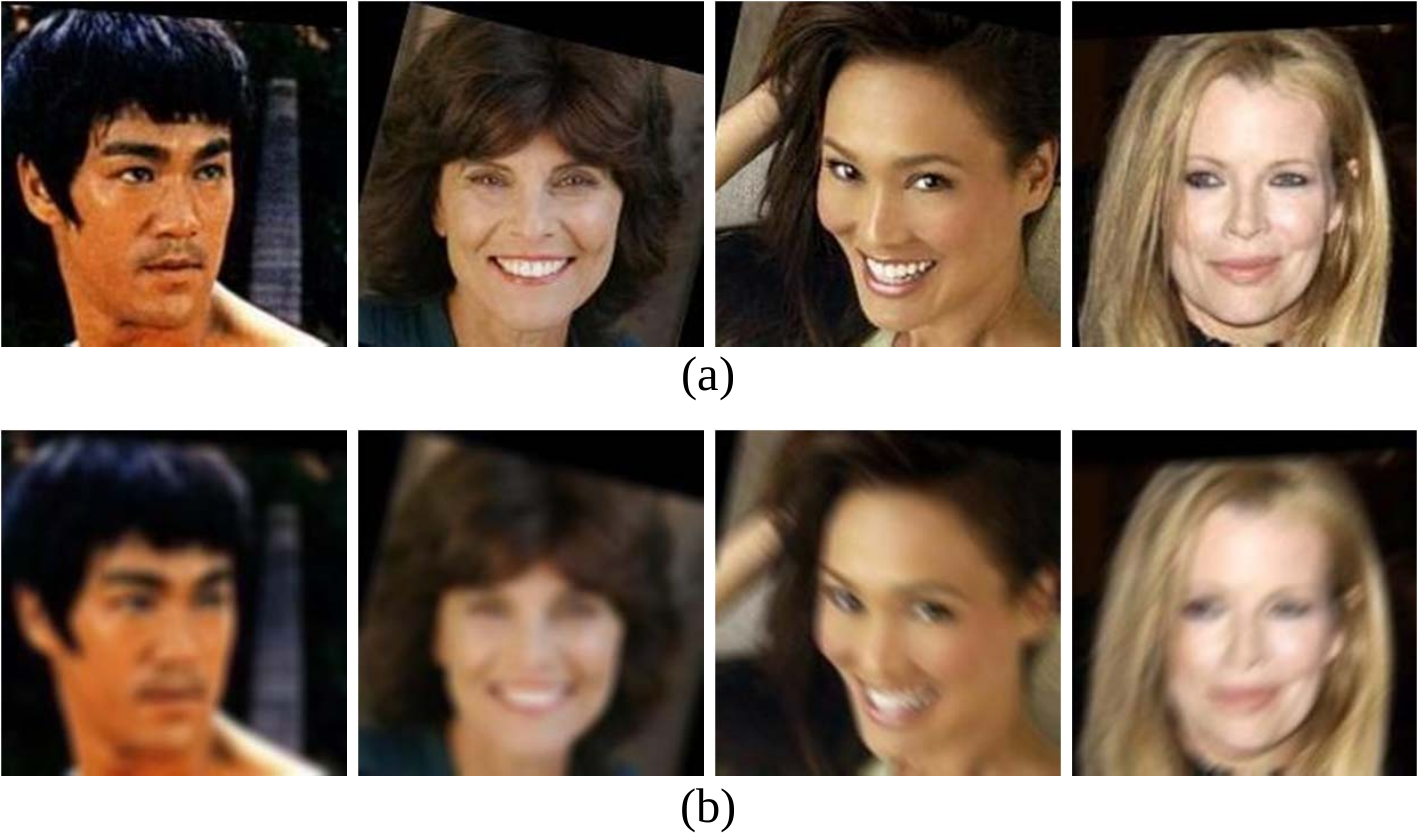}
\caption{Examples of the original still face images and simulated video frames. (a) original still images;
(b) simulated video frames by applying artificial out-of-focus blur (the two figures on the left) and motion blur (the two figures on the right).}
\label{fig:BlurredImages}
\end{figure}

As described above, most available video face databases are rather small and lack diversity in facial variations compared to still face image databases.
We propose artificially generating video-like face data from existing large-scale still face image databases.
Specifically, we simulate two major challenges during surveillance or mobile camera imaging: motion blur and out-of-focus blur.
We assume the blur kernels are uniform since faces usually occupy a small area in video frames.

Due to face movement or mobile device camera shake during exposure, motion blur often appears in video frames.
Supposing the relative motion is along a single direction during exposure,
we can model the motion blur effect by one-dimensional local averaging of neighboring pixels using the following kernel:
\begin{equation}
\label{E:linearmotion}
\mathbf{k}_{m}(i,j;L,\theta)=
\begin{cases}
\frac{1}{L}, &\text{if } \sqrt{i^2+j^2}\leq\frac{L}{2} \text{ and } \frac{i}{j}=-\tan\theta,\\
0, &\text{otherwise},
\end{cases}
\end{equation}
where $(i,j)$ is the pixel coordinate originating from the central pixel,
$L$ is the size of the kernel and indicates the motion distance during exposure, and $\theta$ denotes the motion direction~\cite{wang2014recent}.

Due to the limited depth of field (DOF) of cameras and the large motion range of faces in videos,
out-of-focus blur also occurs in video frames and can be simulated using a uniform kernel~\cite{wang2014recent} or a Gaussian kernel~\cite{nishiyama2011facial}.
In this paper, we employ the Gaussian kernel in the following form:
\begin{equation}
\label{E:outoffocus}
\mathbf{k}_{o}(i,j)=
\begin{cases}
C\cdot\exp\left(-\frac{i^2+j^2}{2\sigma^2}\right),& \text{if }{i}\leq\frac{R}{2} \text{ and } {j}\leq\frac{R}{2},\\
0, & \text{otherwise},
\end{cases}
\end{equation}
where $\sigma$ and $R$ denote the magnitude and size of the kernel, respectively.
$C$ is a constant which ensures the kernel has a unit volume.

Given one still face image $\mathbf{I}_{s}$ and one blur kernel $\mathbf{k}$, the simulated video frame $\mathbf{I}_{v}$ can be obtained by convolution:
\begin{equation}
\label{E:simpleform}
\mathbf{I}_{v}=\mathbf{I}_{s}*\mathbf{k}.
\end{equation}

Based on the above description, we blur each still face image using one randomly sampled blur type from the following 38 possibilities.
Specifically, we choose the value of $L$ from $\{7,9,11\}$ and the value of $\theta$ from $\{0, \pi/4, \pi/2, 3\pi/4\}$;
therefore, there are 12 motion blur choices.
Similarly, we set the value of $R$ as 9 and randomly choose the value of $\sigma$ from $\{1.5,3.0\}$; i.e., there are 2 choices for out-of-focus blur.
We also enrich the blur types by sequentially conducting out-of-focus blur and motion blur, of which there are 24 combinations.
Samples of original still images and the corresponding simulated video frames are illustrated in Fig.~\ref{fig:BlurredImages}.
Since we obtain one blurred image from each still image, we obtain two training data streams of equal size,
i.e., one stream composed of the original still images and the other stream composed of the same number of blurred images.
For CNN training, we provide both training data streams to CNN simultaneously.
Since we encourage each still image and its blurred version to be classified into the same class, CNN automatically learns blur-robust face representations.

It is worth noting that more realistic blur kernels can be found in the literature of deblur.
For example, Sun \textit{et al.}~\cite{sun2015learning} estimated the non-uniform motion blur kernel field from a large field-of-view image.
But it is not clear how to artificially blur images using non-uniform motion blur kernel fields.
Xu \textit{et al.}~\cite{xu2014deep} presented a complicated blur kernel that considers saturation, camera noise, and compression artifacts.
We empirically found that the performance gain obtained by the blur kernel in~\cite{xu2014deep} is slight on the PaSC database compared with the blur kernels in Eq.~\ref{E:linearmotion} and Eq.~\ref{E:outoffocus}. Therefore, we keep using the blur kernels in Eq.~\ref{E:linearmotion} and Eq.~\ref{E:outoffocus} in this paper.

\subsection{Trunk-Branch Ensemble CNN}
\begin{figure*}
\centering
\includegraphics[width=1.0\linewidth]{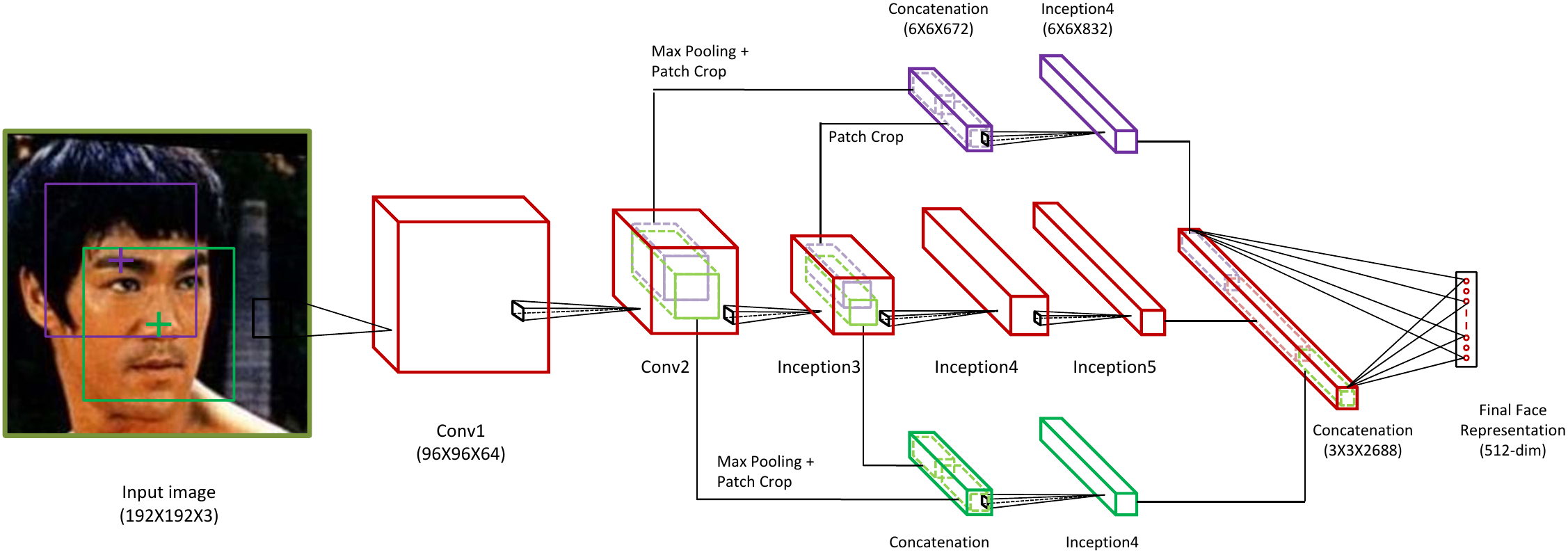}
\caption{Model architecture for Trunk-Branch Ensemble CNN (TBE-CNN).
Note that a max pooling layer is omitted for simplicity following each convolution module, e.g., Conv1 and Inception 3.
TBE-CNN is composed of one trunk network that learns representations for holistic face images
and two branch networks that learn representations for image patches cropped around facial components.
The trunk network and the branch networks share the same low- and middle-level layers,
and they have individual high-level layers.
The output feature maps of the trunk network and branch networks are fused by concatenation.
The output of the last fully connected layer is utilized as the final face representation of one video frame.
}
\label{fig:TBECNN}
\end{figure*}

We propose the Trunk-Branch Ensemble CNN (TBE-CNN) model to efficiently learn pose- and occlusion-robust face representations.
TBE-CNN incorporates one trunk network and several branch networks.
The trunk network is trained to learn face representations for holistic face images,
and each branch network is trained to learn face representations for image patches cropped from one facial component.
In this paper, the trunk network implementation is based on GoogLeNet~\cite{szegedy2014going}.
The most important parameters of the trunk network are tabulated in Table~\ref{tab:GoogLeNet}.
For the other model parameters, we directly follow its original configuration~\cite{szegedy2014going}.
As shown in Table~\ref{tab:GoogLeNet}, we divide the GoogLeNet layers into three levels:
the low-level layers, middle-level layers, and high-level layers.
The three layer levels successively extract features from the low- to the high-level.
Since low- and middle-level features represent local information, the trunk network and branch networks can share low- and middle-level layers.
In comparison, high-level features represent abstract and global information; therefore, different models should have independent high-level layers.
Based on the above observations, the TBE-CNN architecture is illustrated in Fig.~\ref{fig:TBECNN}.

\begin{table}[!t]
\renewcommand{\arraystretch}{1.3}
\caption{Trunk Network Parameters (GoogLeNet)}
\label{tab:GoogLeNet}
\centering
\begin{tabular}{|c|l|c|c|c|}
\hline
&Type (Name)        &\tabincell{c}{Kernel Size/\\Stride}    &Output Size   &Depth\\\hline\hline
\multirow{4}{*}{\rotatebox{90}{Low-Level}}
&convolution (Conv1) &7$\times$7/2         &96$\times$96$\times$64       &1\\\cline{2-5}
&max pool     &2$\times$2/2         &48$\times$48$\times$64       &0\\\cline{2-5}
&convolution (Conv2) &3$\times$3/1         &48$\times$48$\times$192      &2\\\cline{2-5}
&max pool     &2$\times$2/2         &24$\times$24$\times$192      &0\\\hline
\multirow{3}{*}{\rotatebox{90}{Middle-L}}
&inception (3a)     &-              &24$\times$24$\times$256      &2\\\cline{2-5}
&inception (3b)     &-              &24$\times$24$\times$480      &2\\\cline{2-5}
&max pool     &2$\times$2/2         &12$\times$12$\times$480      &0\\\hline
\multirow{9}{*}{\rotatebox{90}{High-Level}}
&inception (4a)     &-              &12$\times$12$\times$512      &2\\\cline{2-5}
&inception (4b)     &-              &12$\times$12$\times$512      &2\\\cline{2-5}
&inception (4c)     &-              &12$\times$12$\times$512      &2\\\cline{2-5}
&inception (4d)     &-              &12$\times$12$\times$528      &2\\\cline{2-5}
&inception (4e)     &-              &12$\times$12$\times$832      &2\\\cline{2-5}
&max pool     &2$\times$2/2         &6$\times$6$\times$832        &0\\\cline{2-5}
&inception (5a)     &-              &6$\times$6$\times$832        &2\\\cline{2-5}
&inception (5b)     &-              &6$\times$6$\times$1024       &2\\\cline{2-5}
&max pool           &2$\times$2/2   &3$\times$3$\times$1024       &1\\\hline
\multirow{2}{*}
&Dropout (0.4)      &-              &3$\times$3$\times$1024       &1\\\cline{2-5}
&Fully-connected    &-              &512                          &1\\\hline
\end{tabular}
\end{table}

The trunk network extracts features from raw pixels.
On top of Inception 5 of the trunk network, we reduce the size of its feature maps to 3$\times$3$\times$1024 by max pooling.
For each branch network,
we directly crop feature maps from Conv2 and Inception 3 module outputs of the trunk network instead of computing low- and middle-level features from scratch.
The cropping size and position are propositional to those of the patches of interest from the input image, as illustrated in Fig.~\ref{fig:TBECNN}.
The size of the feature maps cropped from the Conv2 output is reduced by a half by max pooling
and then concatenated with the feature maps cropped from the Inception 3 output.
The concatenated feature maps form the input of the branch network.
To promote efficiency, each branch network includes only one Inception 4 module.
Similar to the trunk network, the size of feature maps of the branch network is reduced to 3$\times$3$\times$832 by max pooling.

We include two branch networks in TBE-CNN, as illustrated in Fig.~\ref{fig:TBECNN}.
The output feature maps of the trunk network and branch networks are fused by concatenation to form an over-complete face representation,
whose dimension is reduced to 512 by one fully connected layer.
The 512-dimensional feature vector is utilized as the final face representation of one video frame.

\section{TBE-CNN Training}
\label{Sec:Training}
We propose a stage-wise training strategy to effectively optimize the TBE-CNN model parameters.
In the first stage, the trunk network illustrated in Table~\ref{tab:GoogLeNet} is trained alone using softmax loss.
In the second stage, the trunk network parameters are fixed, and each of the branch networks is trained with softmax loss\footnote{
We include one 256-dimensional fully connected layer after the Inception 4 module of each branch network when training it alone.}.
After the trunk network and all branch networks are pre-trained, we fine-tune the complete model illustrated in Fig.~\ref{fig:TBECNN} with softmax loss to fuse the trunk and branch networks.
Finally, to enhance the discriminative power of learnt face representations,
we propose a novel deep metric learning method called Mean Distance Regularized Triplet Loss (MDR-TL) to fine-tune the complete network.

The softmax loss has the advantage of convergence efficiency, but the penalty is imposed on the classification accuracy on training data rather than the discriminative power of learnt face representations.
In comparison, the convergence rate of MDR-TL or the triplet loss is slower, but they directly optimize the discriminative power of the face representations.
In the following, we first briefly review the triplet loss for CNN training~\cite{hoffer2015deep,schroff2015facenet}. Then, we introduce the principle of MDR-TL.
For detailed experimental setting about MDR-TL and the triplet loss, e.g., triplet sampling method and batch size, one can refer to Section 6.3.

\subsection{Brief Review of the Triplet Loss}
Schroff \textit{et al.}~\cite{schroff2015facenet} proposed to employ the triplet loss to train CNNs for face recognition.
In~\cite{schroff2015facenet}, the representation of a face image is $\ell_2$-normalized as the input of the triplet loss function;
therefore, the input face representations for the triplet loss lie on a unit hypersphere.
We denote the $\ell_2$-normalized face representation of one image $\mathbf{x}$ as $\mathbf{f(x)} \in \mathbb{R}^{d}$.
For accurate face recognition, it is desired that the representation of an image $\mathbf{f(x}^a)$ (anchor) of a specific subject is closer to each image representation $\mathbf{f(x}^p)$ (positive)
of the same subject than it is to any image representation $\mathbf{f(x}^n)$ (negative) of other subjects.
$\{\mathbf{f(x}^a),\mathbf{f(x}^p),\mathbf{f(x}^n)\}$ compose a triplet.
Formally, each triplet is desired to satisfy the triplet constraint~\cite{schroff2015facenet}:
\begin{align}\label{E:tlConstraint}
\|\mathbf{f(x}^a)-\mathbf{f(x}^p)\|_2^2 + \beta < \|\mathbf{f(x}^a)-\mathbf{f(x}^n)\|_2^2,
\end{align}
where $\beta$ is the margin that is enforced between the positive pair $\{\mathbf{f(x}^a),\mathbf{f(x}^p)\}$ and the negative pair $\{\mathbf{f(x}^a),\mathbf{f(x}^n)\}$.
The triplet loss function is formulated as $L_{triplet}(\mathbf{f})=$
\begin{align}\label{E:triplet}
\frac{1}{2N} \sum_{i=1}^N \left[\|\mathbf{f(x}_{i}^a)-\mathbf{f(x}_{i}^p)\|_2^2-\|\mathbf{f(x}_{i}^a)-\mathbf{f(x}_{i}^n)\|_2^2+\beta \right]_{+},
\end{align}
where $N$ is the number of triplets that violate the triplet constraint in a batch
and $\{\mathbf{f(x}_{i}^a),\mathbf{f(x}_{i}^p),\mathbf{f(x}_{i}^n)\}$ stands for the $i$-th triplet.
In this paper, triplets are sampled online within each batch.

\subsection{Mean Distance Regularized Triplet Loss}
\begin{figure}
\centering
\includegraphics[width=1.0\linewidth]{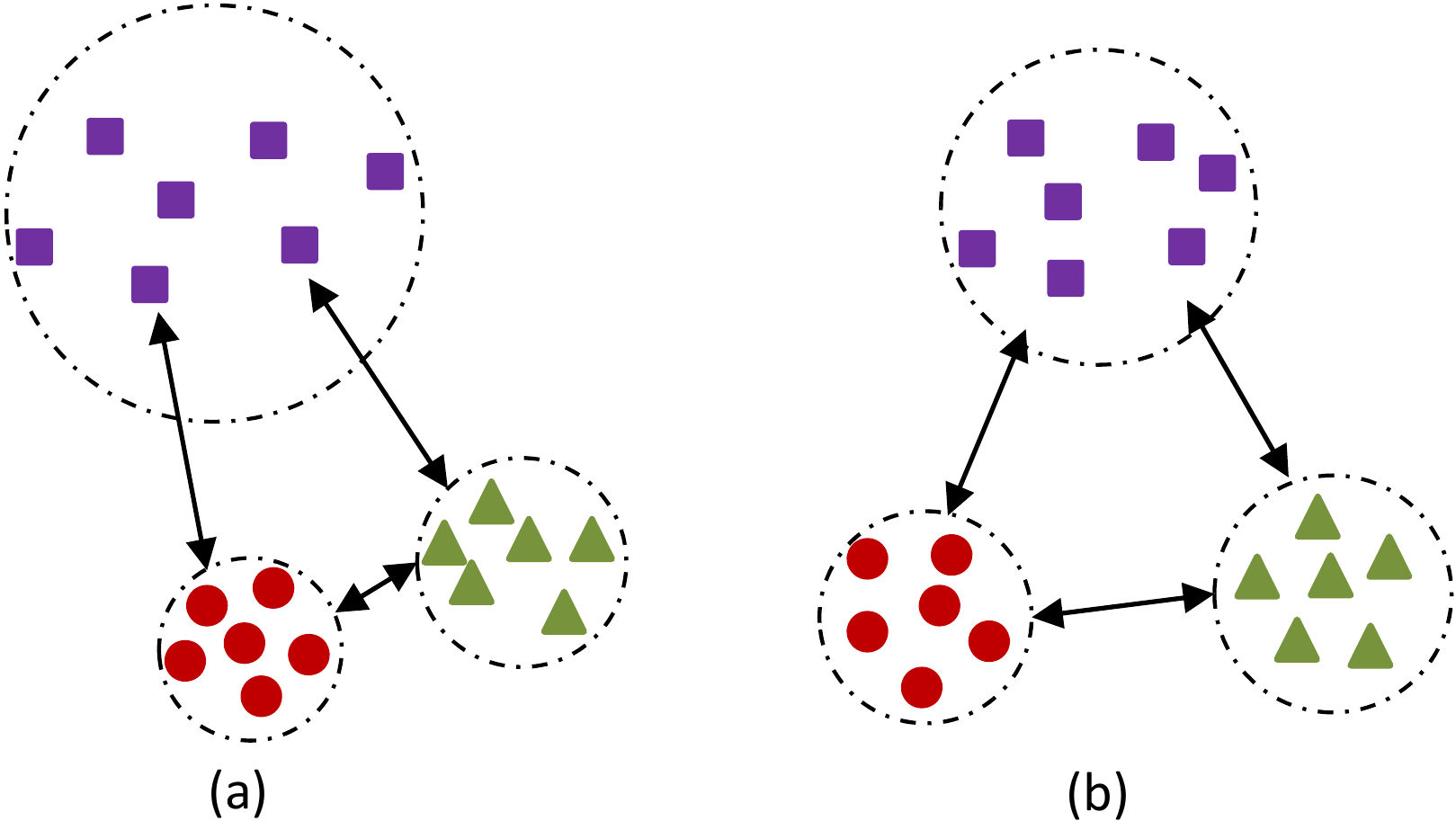}
\caption{The principle of Mean Distance Regularized Triplet Loss (MDR-TL).
(a) Triplets sampled in the training batch satisfy the triplet constraint (Eq.~\ref{E:tlConstraint}).
However, due to the non-uniform intra- and inter-class distance distributions,
it is hard to select an ideal threshold for face verification.
(b) MDR-TL regularizes the triplet loss by setting a margin for the distance between subject mean representations.}
\label{fig:MDRTL}
\end{figure}

Both the pairwise loss~\cite{sun2014deep} and the triplet loss~\cite{hoffer2015deep,schroff2015facenet} rely on sampling effective image pairs or triplets in a batch from possible combinations.
Take the triplet loss for example, the optimization of $L_{triplet}(\mathbf{f})$ is based on each individual triplet $\{\mathbf{f(x}_{i}^a),\mathbf{f(x}_{i}^p),\mathbf{f(x}_{i}^n)\}(1\leq{i}\leq{N})$ in Eq.~\ref{E:triplet}.
However, the inter- and intra-class distance distributions are not explicitly constrained, which has a negative impact on face recognition.
For example, the training samples in Fig.~\ref{fig:MDRTL}(a) satisfy the triplet constraint in Eq.~\ref{E:tlConstraint};
but due to the non-uniform inter- and intra-class distance distributions,
it is difficult to find an ideal threshold in face verification to determine whether a pair of face images belong to the same subject or not.
To overcome this problem, we propose the MDR-TL loss function,
which regularizes the triplet loss by constraining the distance between subject mean representations.

\begin{figure}
\centering
\includegraphics[width=1.0\linewidth]{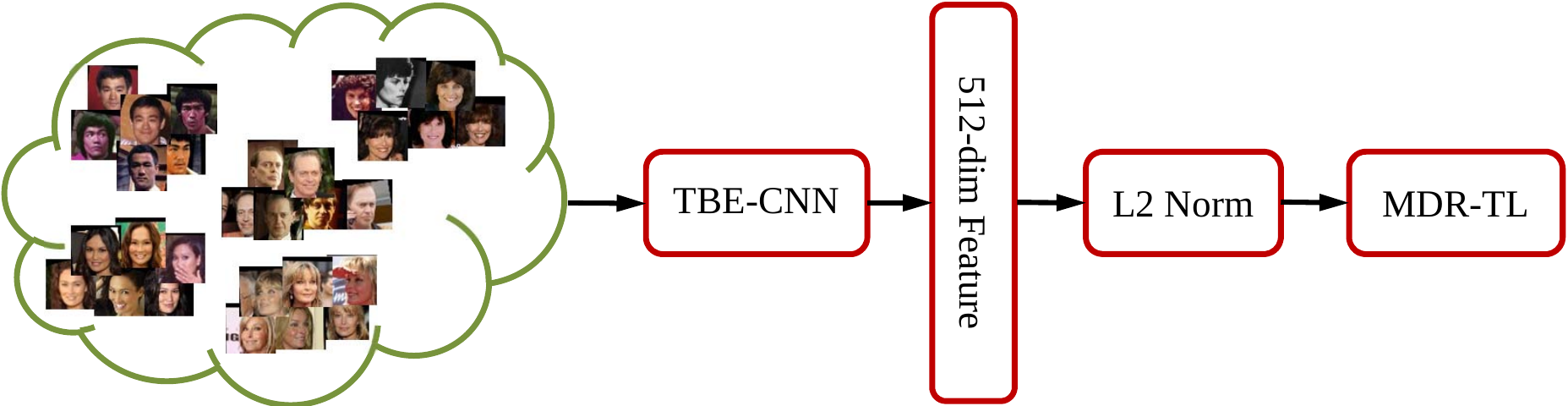}
\caption{Illustration of TBE-CNN training with MDR-TL.
MDR-TL is employed to further enhance the discriminative power of learnt face representations.}
\label{fig:TBE_MDR}
\end{figure}

Similar to~\cite{schroff2015facenet}, the 512-dimensional face representation extracted by TBE-CNN is $\ell_2$-normalized as the MDR-TL input, as illustrated in Fig.~\ref{fig:TBE_MDR}.
Two constraints are included in MDR-TL.
In the first, the mean representations of different subjects should be well separated to ensure that the inter-class distance distribution is uniform.
We realize this constraint by enforcing a margin $\alpha$ between the mean representation $\hat{\bm{\mu}}_{c}$ of one subject and its nearest mean representation $\hat{\bm{\mu}}_{c}^n$ in the same batch:
\begin{equation}\label{E:InterConstraint}
\begin{array}{ll}
\|\hat{\bm{\mu}}_{c}-\hat{\bm{\mu}}_{c}^n\|_2^2>\alpha,
\end{array}
\end{equation}
where $\hat{\bm{\mu}}_{c}$ is the estimated mean representation for the $c$-th subject that also lies on the unit hypersphere. For simplicity, we compute $\hat{\bm{\mu}}_{c}$ in the following way:
\begin{align}
\hat{\bm{\mu}}_{c}=\frac{\bm{\mu}_{c}}{\|\bm{\mu}_{c}\|},
\end{align}
and
\begin{align}
\bm{\mu}_{c}=\frac{1}{N_{c}} \sum_{i=1}^{N_{c}}\mathbf{f(x}_{ci}),
\end{align}
where $N_{c}$ is the number of images of the $c$-th subject in the current batch and $\mathbf{f(x}_{ci})$ denotes the face representation of the $i$-th image for the $c$-th subject.
The nearest mean representation $\hat{\bm{\mu}}_{c}^n$ for $\hat{\bm{\mu}}_{c}$ is detected online efficiently within the same batch.

In the second, triplets sampled in a training batch should satisfy the triplet constraint in Eq.~\ref{E:tlConstraint}.
Working together with the first constraint, the triplet constraint is helpful to regularize the intra-class distance distribution to be uniform.
Based on the above analysis, we formulate MDR-TL as the following optimization problem:
\begin{equation}\label{E:MDRTL}
\min\limits_{\mathbf{f}}L(\mathbf{f})=L_{triplet}(\mathbf{f}) + L_{mean}(\mathbf{f}),
\end{equation}
where
\begin{align}\label{E:meanDistance}
L_{mean}(\mathbf{f})=\frac{1}{2P} \sum_{c=1}^C \max(0, \alpha-\|\hat{\bm{\mu}}_{c}-\hat{\bm{\mu}}_{c}^n\|_2^2).
\end{align}

In Eq.~\ref{E:MDRTL}, we assume equal weights for $L_{triplet}(\mathbf{f})$ and $L_{mean}(\mathbf{f})$, which empirically works well.
In Eq.~\ref{E:meanDistance}, $C$ is the number of subjects in the current batch,
and $P$ is the number of mean representations that violate the constraint in Eq.~\ref{E:InterConstraint}.
We optimize Eq.~\ref{E:MDRTL} using the standard stochastic gradient descent with momentum~\cite{jia2014caffe}.
The gradient of $L$ with respect to $\mathbf{f(x}_{ci})$ is derived as follows,
\begin{align}
\frac{\partial L}{\partial \mathbf{f(x}_{ci})} = \frac{\partial L_{triplet}}{\partial \mathbf{f(x}_{ci})} + \frac{\partial L_{mean}}{\partial \mathbf{f(x}_{ci})},
\end{align}
where
\begin{align}
\frac{\partial L_{mean}}{\partial \mathbf{f(x}_{ci})} = -\frac{1}{P} (\sum_{j=1,j\neq c}^{C}w_{j}(\hat{\bm{\mu}}_{c}-\hat{\bm{\mu}}_{j})^{T}) \frac{\partial \hat{\bm{\mu}}_{c}}{\partial \mathbf{f(x}_{ci})}.
\end{align}
$w_{j}$ equals 1 if the constraint in Eq.~\ref{E:InterConstraint} is violated,
and $\hat{\bm{\mu}}_{j}$ is the nearest neighbor of $\hat{\bm{\mu}}_{c}$ or $\hat{\bm{\mu}}_{c}$ is the nearest neighbor of $\hat{\bm{\mu}}_{j}$.
Otherwise, $w_{j}$ equals 0.

\begin{align}
\notag \frac{\partial \hat{\bm{\mu}}_{c}}{\partial \mathbf{f(x}_{ci})} &= \|\bm{\mu}_{c}\|^{-1}\frac{\partial \bm{\mu}_{c}}{\partial \mathbf{f(x}_{ci})} + \bm{\mu}_{c}\frac{\partial \|\bm{\mu}_{c}\|^{-1}}{\partial \mathbf{f(x}_{ci})}\\
\notag &=\frac{1}{N_{c}\|\bm{\mu}_{c}\|} (\mathbf{I}-\hat{\bm{\mu}}_{c}\hat{\bm{\mu}}_{c}^{T}), \\
\end{align}
and can be approximated by $\frac{\mathbf{I}}{N_{c}\|\bm{\mu}_{c}\|}$ in practice.
The derivative of $L$ with respect to model parameters can be computed via the chain rule.
As illustrated in Fig.~\ref{fig:MDRTL}(b), by setting a margin between the mean representations of different subjects,
the inter- and intra-class distance distributions become uniform.
This regularization is empirically verified as important below.
Another possible regularization term for MDR-TL can be constructed by explicitly controlling the variance of intra-class distances for each subject,
but no performance promotion compared with MDR-TL is observed in our experiments.

\section{VFR with TBE-CNN}
\label{Sec:VFR}
We next employ the TBE-CNN model for VFR.
Given a video, we first augment its video frames by horizontal flipping.
Then, all video frames pass through the TBE-CNN network, and fusing their outputs by average pooling produces the compact video representation.
The representation of one still image can be extracted in a similar way by assuming it is a single-frame video.
For all three settings - S2V, V2S, and V2V matching - we consistently employ the cosine metric to calculate the similarity $\rho$ between the two representations $\mathbf{y}_{1}$ and $\mathbf{y}_{2}$:

\begin{align}
\rho\left(\mathbf{y}_{1}, \mathbf{y}_{2}\right) = \frac{\mathbf{y}_{1}^{T}\mathbf{y}_{2}}{\|\mathbf{y}_{1}\|\|\mathbf{y}_{2}\|}.
\end{align}

\section{EXPERIMENTS}
\label{Sec:EXPERIMENTS}
\begin{figure*}
\centering
\includegraphics[width=0.99\linewidth]{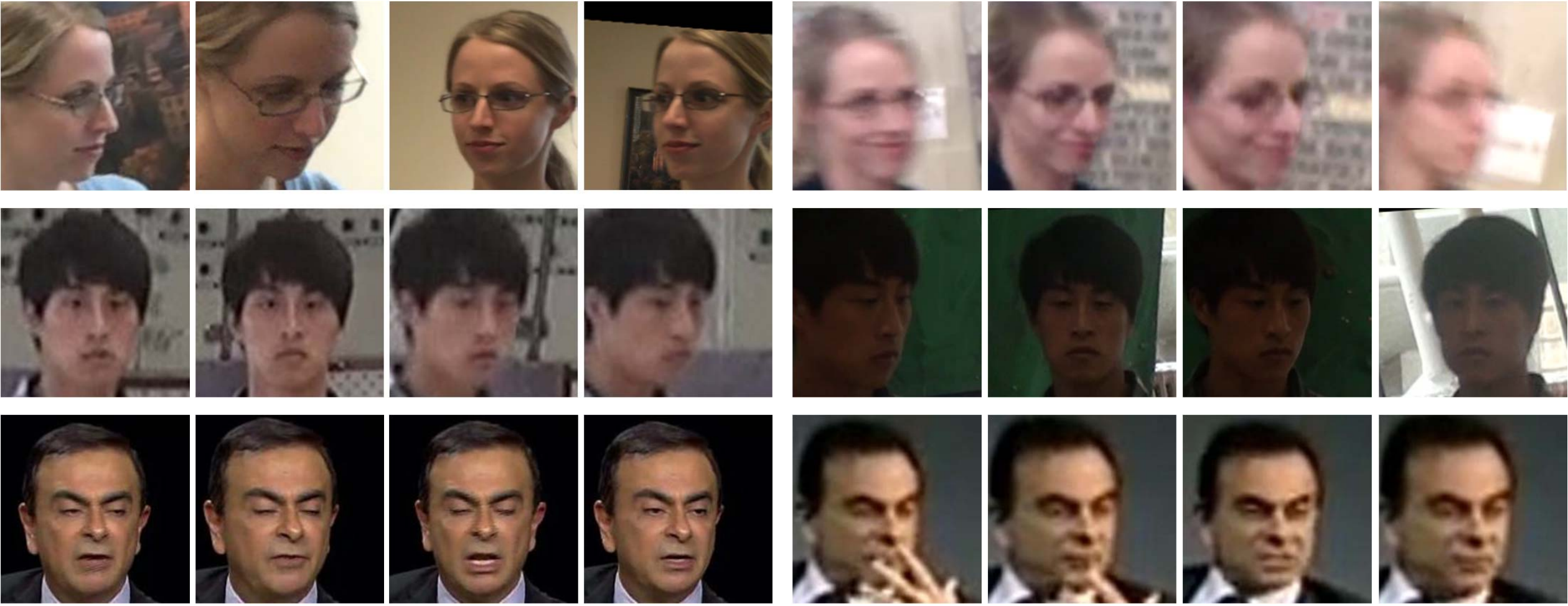}
\caption{Sample video frames after normalization: PaSC (first row), COX Face (second row), and YouTube Faces (third row).
For each database, the four frames on the left are sampled from a video recorded under relatively good conditions,
and the four frames on the right are selected from low-quality video.
}
\label{fig:Database}
\end{figure*}

We now systematically evaluate the proposed TBE-CNN framework for VFR.
Experiments are conducted on three publicly available large-scale video face databases: PaSC, COX Face, and YouTube Faces.
Example images are shown in Fig.~\ref{fig:Database}.

The PaSC database~\cite{beveridge2013challenge} contains 2,802 videos of 265 subjects.
The videos were recorded by multiple sensors in varied indoor and outdoor locations.
They are divided into two sets: a control set and a handheld set.
High-end cameras installed on tripods captured videos in the control set to produce images of good quality,
as illustrated by the first four images in the first row in Fig.~\ref{fig:Database}.
Videos in the handheld set were captured by five handheld video cameras.
For each video, the subject was asked to carry out actions to create a wide range of poses at variable distances from the camera.
Faces in the video exhibit serious motion and out-of-focus blur and rich pose variations.
Here, we adopted the officially defined V2V matching protocol for face verification~\cite{beveridge2013challenge}.

The COX Face database~\cite{Huang2015Benchmark} incorporates 1,000 still images and 3,000 videos of 1,000 subjects.
A high-quality camera in well-controlled conditions captured still images to simulate ID photos.
The videos were taken while the subjects were walking in a large gym to simulate surveillance.
Three cameras at different locations were installed to capture videos of the walking subject simultaneously.
Videos captured by the three cameras created three subsets: Cam1, Cam2, and Cam3.
Since COX Face has significantly more subjects compared to PaSC, it is an ideal database for face identification.
The standard S2V, V2S, and V2V matching protocols~\cite{Huang2015Benchmark} for face identification were adopted.

The YouTube Faces database~\cite{wolf2011face} includes 3,425 videos of 1,595 subjects.
All videos were downloaded from the YouTube website.
Since the majority of subjects in this database were in interviews, there is no obvious image blur or pose variation.
Instead, this database is low resolution and contains serious compression artifacts, as shown in the last four video frames of the third row in Fig.~\ref{fig:Database}.
5,000 video pairs were collected from the videos, which were divided into 10 equally sized splits.
Each split contained 250 homogeneous video pairs and 250 heterogeneous video pairs.
Following the ``restricted'' protocol defined in~\cite{wolf2011face}, we report the mean verification accuracy and the standard error of the mean ($S_{E}$) on the 10 splits.

For all three video databases, we employ the face detection results provided by the respective databases.
There are 60 videos in PaSC that have no face being detected. We ignore these 60 videos since their similarity score to any video is NaN,
and thus have no impact on the verification rate or Receiver Operating Characteristic (ROC) curve.
Each detected face is normalized and resized to $192\times192$ pixels using an affine transformation based on the five facial feature points detected by~\cite{sun2013deep}.
Sample face images after normalization are shown in Fig.~\ref{fig:Database}.
A number of experiments are conducted. First, the effectiveness of video training data simulation for CNN is tested.
Second, the MDR-TL performance is evaluated.
Third, the trunk network performance and TBE-CNN are compared.
Finally, the performance of the complete TBE-CNN framework is compared to state-of-the-art VFR methods on the PaSC, COX Face, and YouTube Faces databases.

\subsection{Implementation Details of TBE-CNN}
The proposed TBE-CNN model is trained on the publicly available CASIA-WebFace database~\cite{yi2014learning} and deployed in experiments on the three video face databases described above.
The CASIA-WebFace database contains 494,414 images of 10,575 subjects.
We augment the CASIA-WebFace database by horizontal flipping and image jittering~\cite{ding2015robust}.
After augmentation, the size of the training set is about 2.68 million, which forms the still image training data stream.
According to the description in Section~\ref{Sec:TBECNN}, we blur each still image using a randomly chosen kernel to form the simulated video data stream.

During training, the batch size is set to 90 for both the still image stream and simulated video stream to produce a complete batch size of 180.
Following the stage-wise training framework introduced in Section~\ref{Sec:Training},
the trunk network is trained for 13 epochs using softmax loss, with the learning rate gradually decreased from 0.01 to 0.001.
Then, the branch networks are trained for 4 epochs.
Next, the complete TBE-CNN model is fine-tuned for another 4 epochs with softmax loss to fuse the trunk and branch networks with a small learning rate of 0.001.
Finally, the TBE-CNN model is fine-tuned with MDR-TL for one more epoch with the learning rate of 0.001.
The open-source deep learning package Caffe~\cite{jia2014caffe} is utilized to train the deep models.

The first three experiments are conducted on PaSC. For these three experiments,
we manually remove the falsely detected faces from the face detection results provided by the database\footnote{The defect in face detection is because several different faces appeared in the same video. The list of manually removed faces is available upon request.}
to help us to accurately reflect the effectiveness of proposed algorithms.
For the fourth experiment, we provide results based on both the original face detections and manually corrected face detections.

\subsection{Effectiveness of Simulated Video Training Data}
This experiment provides evidence to justify the proposed video data simulation strategy.
The evaluation is based on the trunk network with softmax loss.
Four types of training data are evaluated: the still image training data stream alone (SI), the simulated video training data stream alone (SV),
the fine-tuning (FT) strategy adopted for VFR~\cite{Ross2015report},
and the two-stream training data (TS) that combines both SI and SV.
For SI, SV, and TS, the networks are trained with the same number of iterations (13 epochs),
and the amount of augmented training data is 2.6 million, 2.6 million, and 5.2 million, respectively.
For FT, we fine-tune the network produced by SI with real-world video data.
Following~\cite{Ross2015report}, we use all video frames in the COX Face database and the PaSC training set for fine-tuning.
The two video sets incorporate 0.45 million video frames, comparable in volume to CASIA-WebFace.
For fair comparison with TS, we augment the video frames with horizontal flipping and jittering to 2.6 million.
Therefore, the total amount of training data for FT is 5.2 million.
The learning rate of the fine-tuning stage for FT is set to 0.001, and we observe that its performance saturates on PaSC after one epoch.

\begin{figure}
\centering
\includegraphics[width=1.0\linewidth]{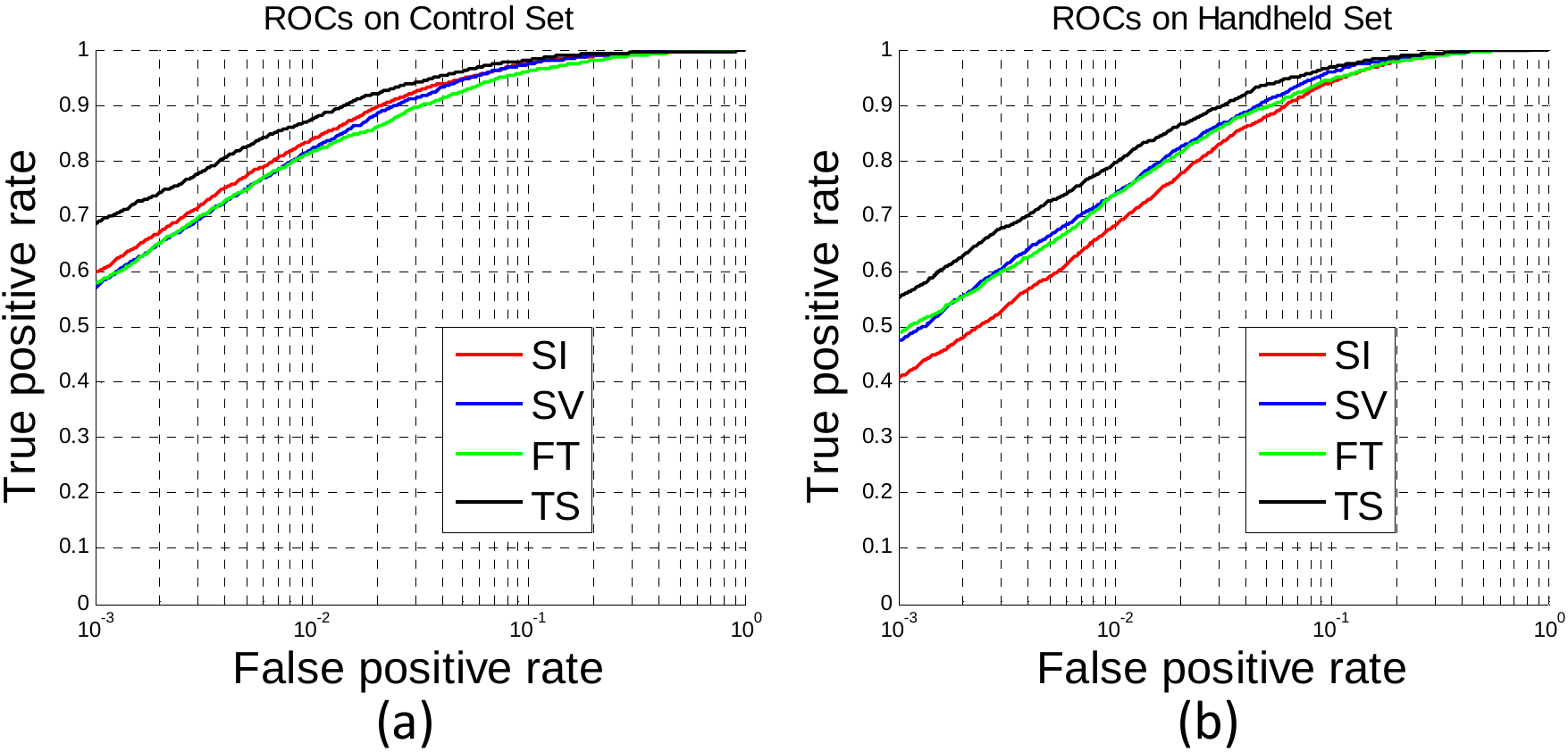}
\caption{ROC curves of the trunk network trained with different types of training data on the PaSC database.
(a) Control set; (b) handheld set.
}
\label{fig:Exp1Roc}
\end{figure}

The verification rates at 1\% False Acceptance Rate (FAR) of the above four types of training data are tabulated in Table~\ref{tab:PaSC_TrainStrategy},
and the corresponding ROC curves are illustrated in Fig.~\ref{fig:Exp1Roc}.
While SI slightly outperforms SV on the control set, SV significantly outperforms SI by as much as 5.6\% on the handheld set.
This is due to the difference in image quality between the control and handheld sets.
Specifically, high-end cameras recorded video frames in the control set; therefore, the image quality is similar to that of the training data for SI.
In comparison, low-quality mobile cameras captured videos in the handheld set and its frames sufferred from severe image blur,
as illustrated in Fig.~\ref{fig:Database}.
The clear performance advantage of SV on the handheld set justifies the use of simulated video training data.

The performance of FT is comparable to SV, but at the cost of more training data and longer training time.
In comparison, TS significantly outperforms SI, SV, and FT on both sets.
Compared to SI and SV, TS combines the training data of both still images and simulated video frames,
which regularizes the model to learn blur-robust representations.
Compared to FT, which has the same size of training data, TS shows a performance advantage of 5.9\% and 5.6\% for the control and handheld sets, respectively.
This is for two main reasons: first, there is only slight adaptation of CNN model parameters by FT to the VFR task compared to TS,
which directly optimizes all model parameters of CNN for VFR;
second, frames in the same video are highly redundant.
Therefore, fine-tuning with a small amount of real-world video data may suffer from over-fitting.
Moreover, another important advantage of TS over FT lies in S2V and V2S tasks, detailed in Section~\ref{sec:cox}.

\begin{table}[!t]
\renewcommand{\arraystretch}{1.3}
\caption{Verification Rates (\%) at 1\% FAR on PaSC with Different Types of Training Data}
\label{tab:PaSC_TrainStrategy}
\centering
\begin{tabular}{|l|c|c|c|c|c|}
\hline
                &\tabincell{c}{\# Training Data\\(Augmented)} &Control Set     &Handheld Set\\\hline\hline
SI              &2.6M    &83.72   &68.20\\\hline
SV              &2.6M    &81.89   &73.80\\\hline
FT              &5.2M    &81.41   &73.70\\\hline
TS              &5.2M    &\textbf{87.31}   &\textbf{79.33}\\\hline
\end{tabular}
\end{table}

\subsection{Effectiveness of MDR-TL}
This experiment evaluates the performance of the proposed MDR-TL loss function.
Two baseline loss functions for deep metric learning are compared: the pairwise loss~\cite{zhang2009patch} and the triplet loss~\cite{schroff2015facenet}.
Similar to the previous experiment, performance comparisons are based on the trunk network.
In the following experiments, we remove the two auxiliary classifiers of the trunk network, i.e., GoogLeNet, since it has been well pre-trained by the softmax loss.
In other words, the metric learning-based loss functions are employed to replace the last classifier of the trunk network~\cite{szegedy2014going}.
Two representative training data types are considered:
1) SI, which is the most widely used training data type; and 2) our proposed TS.
The models produced in the previous experiment are fine-tuned with a learning rate of 0.001 for one epoch with each of the three loss functions.
We observe that the performance saturates after one epoch of fine-tuning.
Image pairs or triplets are sampled within each batch online. To ensure that sufficient image pairs or triplets are sampled,
we re-arrange the training samples such that each subject in a batch contains 6 images.
Therefore, there are 30 subjects in total for a batch size of 180.
For the pairwise loss, our implementation is based on the Discriminative Locality Alignment (DLA) model~\cite{zhang2009patch},
which has three parameters to tune.
We empirically find that sampling all positive pairs and the same number of hard-negative pairs is optimal for DLA.
For the triplet loss, following the specifications in~\cite{schroff2015facenet},
we utilize all positive image pairs and randomly sample semi-hard negative samples to compose triplets.
We traverse the value of the margin parameter $\beta$ within $\{0.1, 0.2, 0.3, 0.4\}$ and report its best performance.
Similarly, we traverse the value of the margin parameter $\alpha$ for MDR-TL within $\{1.6, 1.8, 2.0, 2.2\}$ and report the best result.
It is worth noting that our implementation of the $L_{triplet}(\mathbf{f})$ term in Eq.~\ref{E:MDRTL} for both the triplet loss and MDR-TL is exactly the same.
The optimal values for $\alpha$ and $\beta$ are 2.0 and 0.2, respectively.
The verification rates at 1\% FAR of the three loss functions are compared in Table~\ref{tab:Exp2_MDRTL_SemiHard}.

\begin{table}[!t]
\renewcommand{\arraystretch}{1.3}
\caption{Verification rates (\%) at 1\% FAR with different loss functions on PaSC (Semi-hard Negative Mining)}
\label{tab:Exp2_MDRTL_SemiHard}
\centering
\begin{tabular}{|l|c|c||c|c|}
\hline
\multirow{2}{*}{}  \multirow{2}{*}{} &
\multicolumn{2}{c||}{SI Training Data} &
\multicolumn{2}{c|}{TS Training Data} \\
\cline{2-5}
&\tabincell{c}{Control\\Set} &\tabincell{c}{Handheld\\Set} &\tabincell{c}{Control\\Set} &\tabincell{c}{Handheld\\Set} \\
\hline\hline
Pairwise loss   &94.61   &87.16   &95.35   &92.03\\\hline
Triplet loss    &94.54   &87.00   &96.03   &92.46\\\hline
MDR-TL          &\textbf{95.86}   &\textbf{89.60}   &\textbf{96.66}   &\textbf{93.91}\\\hline
\end{tabular}
\end{table}

\begin{table}[!t]
\renewcommand{\arraystretch}{1.3}
\caption{Verification rates (\%) at 1\% FAR with different loss functions on PaSC (Hard Negative Mining)}
\label{tab:Exp2_MDRTL_Hard}
\centering
\begin{tabular}{|l|c|c||c|c|}
\hline
\multirow{2}{*}{}  \multirow{2}{*}{} &
\multicolumn{2}{c||}{SI Training Data} &
\multicolumn{2}{c|}{TS Training Data} \\
\cline{2-5}
&\tabincell{c}{Control\\Set} &\tabincell{c}{Handheld\\Set} &\tabincell{c}{Control\\Set} &\tabincell{c}{Handheld\\Set} \\
\hline\hline
Triplet loss    &94.78   &86.77   &96.07   &92.73\\\hline
MDR-TL          &\textbf{95.66}   &\textbf{89.47}   &\textbf{96.64}   &\textbf{93.61}\\\hline
\end{tabular}
\end{table}

\begin{table}[!t]
\renewcommand{\arraystretch}{1.3}
\caption{Verification rates (\%) at 1\% FAR with different loss functions on PaSC (Hardest Negative Mining)}
\label{tab:Exp2_MDRTL_Hardest}
\centering
\begin{tabular}{|l|c|c||c|c|}
\hline
\multirow{2}{*}{}  \multirow{2}{*}{} &
\multicolumn{2}{c||}{SI Training Data} &
\multicolumn{2}{c|}{TS Training Data} \\
\cline{2-5}
&\tabincell{c}{Control\\Set} &\tabincell{c}{Handheld\\Set} &\tabincell{c}{Control\\Set} &\tabincell{c}{Handheld\\Set} \\
\hline\hline
Triplet loss    &93.86   &85.35   &95.22   &91.47\\\hline
MDR-TL          &\textbf{95.32}   &\textbf{88.35}   &\textbf{96.00}   &\textbf{93.12}\\\hline
\end{tabular}
\end{table}

\begin{figure}
\centering
\includegraphics[width=1.0\linewidth]{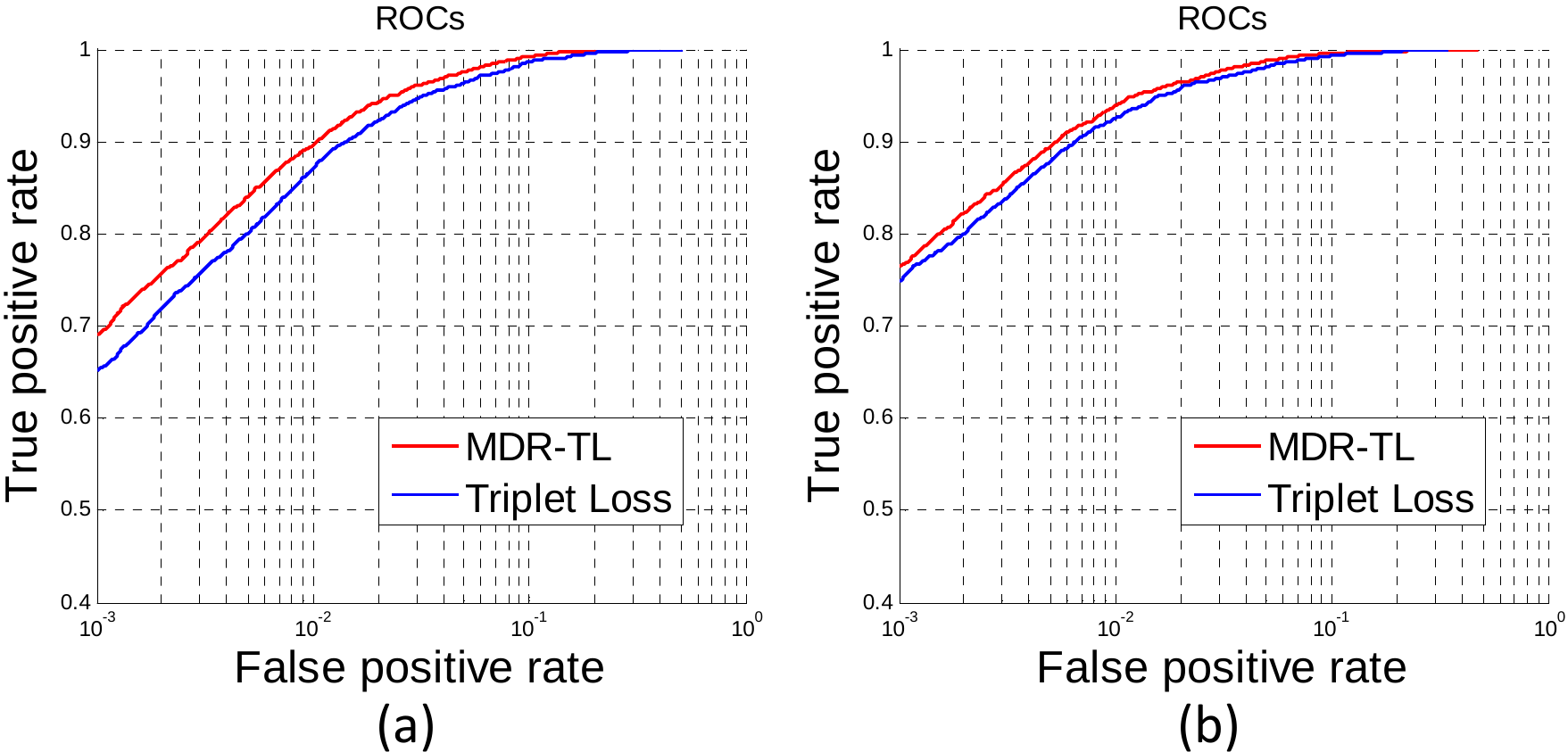}
\caption{ROC curves of MDR-TL and the triplet loss functions on the handheld set of PaSC.
(a) SI training data; (b) TS training data.}
\label{fig:Exp2_MDRTL_Roc}
\end{figure}

It is shown that the performance of the pairwise loss (DLA) and the triplet loss are comparable with the SI training data.
But the triplet loss visibly outperforms DLA with the TS training data.
One major disadvantage of DLA is that it has more parameters which substantially affect the performance.
In comparison, the triplet loss has only one parameter and its optimal value is stable across experiments.
MDR-TL outperforms the baselines consistently under all settings.
In particular, for the most challenging setting, i.e., the handheld set of PaSC with SI training data, MDR-TL outperforms the triplet loss by as much as 2.6\%.
With TS training data, the performance margin between MDR-TL and the triplet loss is around 1\%, which means about 20\% reduction in verification error rate.
The ROC curves of the two loss functions on the handheld set of PaSC are plotted in Fig.~\ref{fig:Exp2_MDRTL_Roc}.
The results of experiments justify the role of $L_{mean}(\mathbf{f})$ in Eq.~\ref{E:MDRTL} as an effective regularization term to the triplet loss.

We also compared the performance of the triplet loss and MDR-TL with another two popular negative sampling strategies,
i.e., hard negative sampling~\cite{schroff2015facenet} and hardest negative sampling~\cite{simo2015discriminative}.
Results of experiments are tabulated in Table~\ref{tab:Exp2_MDRTL_Hard} and Table~\ref{tab:Exp2_MDRTL_Hardest}, respectively.
It is clear that MDR-TL outperforms the triplet loss under all settings.
It is also shown that the hardest negative sampling method is inferior to semi-hard negative mining and hard negative mining methods.
Besides, the performance of hard negative mining and semi-hard negative mining is equivalent.
This is because that the training set, i.e., CASIA-WebFace, was collected in a semi-automatic way;
therefore, the labels of the training images are noisy.
In this case, selecting the hardest negative sample is suboptimal due to the sensitivity to the labeling error.

The comparisons in Table~\ref{tab:Exp2_MDRTL_SemiHard}, Table~\ref{tab:Exp2_MDRTL_Hard}, Table~\ref{tab:Exp2_MDRTL_Hardest}, and Fig.~\ref{fig:Exp2_MDRTL_Roc}
also convincingly show that the proposed TS strategy to compose CNN training data is essential to achieve blur-robustness in VFR.
Briefly, for the control and handheld sets of PaSC, CNN performance with TS training data is higher than that with SI training data by around 1.0\% and 5.0\%, respectively.

\subsection{Effectiveness of Trunk-Branch Fusion}
\begin{figure}
\centering
\includegraphics[width=0.75\linewidth]{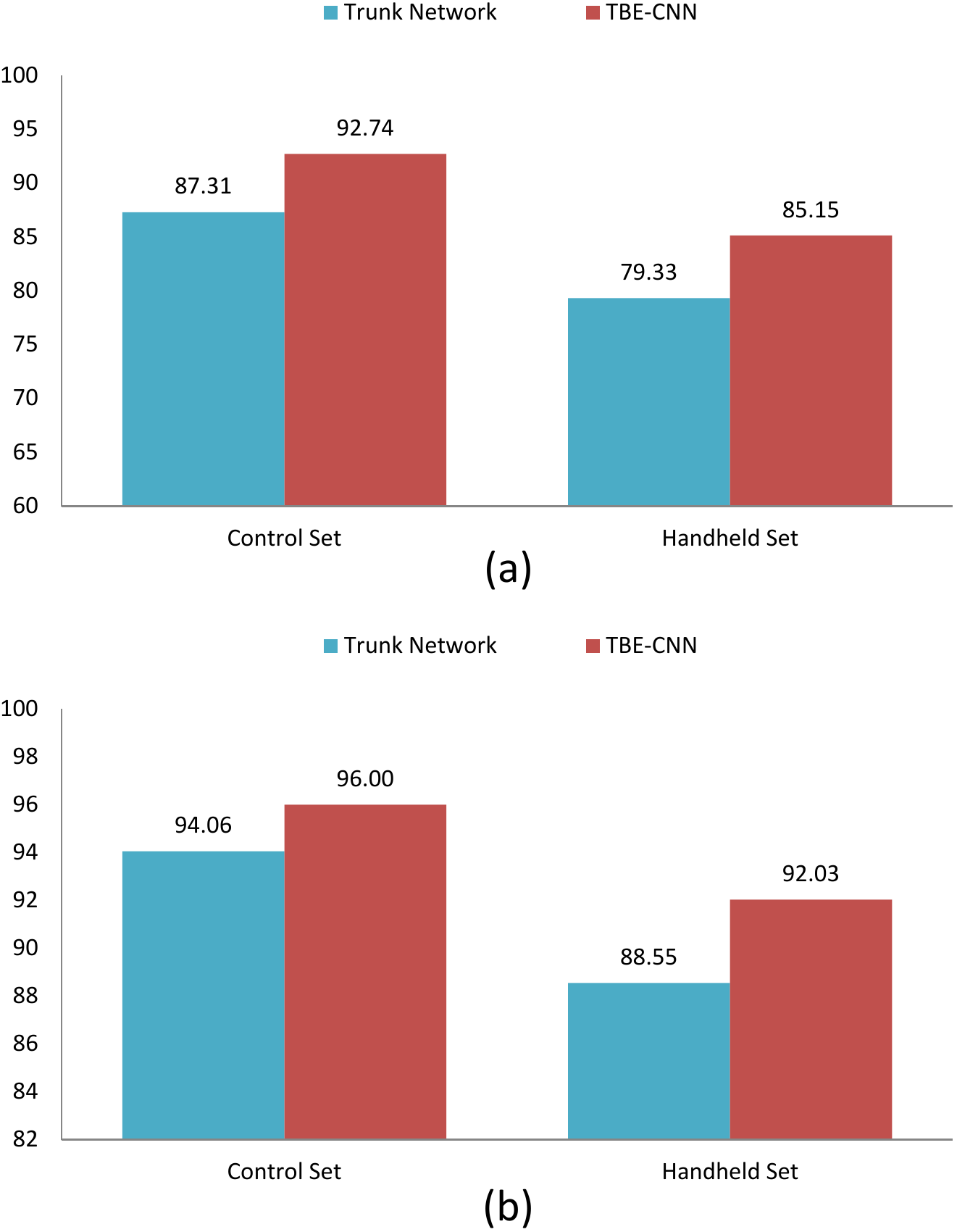}
\caption{Verification rates (\%) at 1\% FAR by the trunk network and TBE-CNN. Comparison is based on the softmax loss.
(a) Performance comparison without BN layers; (b) performance comparison with BN layers.}
\label{fig:Exp3_Fusion}
\end{figure}

\begin{figure}
\centering
\includegraphics[width=1.0\linewidth]{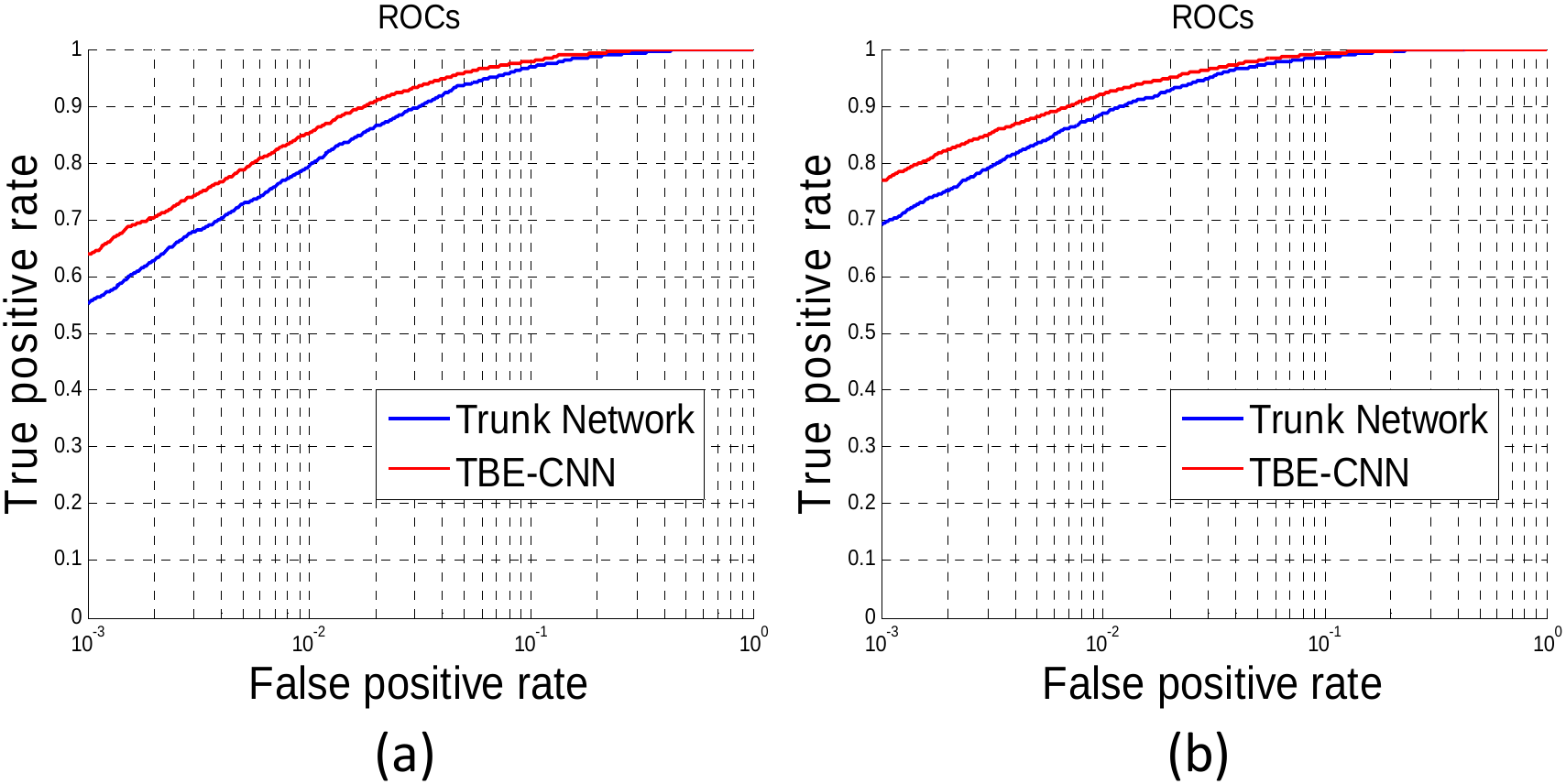}
\caption{ROC curves of the trunk network and TBE-CNN on the handheld set of PaSC. (a) Without BN layers; (b) with BN layers.}
\label{fig:Exp3_TBECNNvsTrunk}
\end{figure}

The verification rates at 1\% FAR of TBE-CNN and trunk network with TS training data on PaSC are compared in Fig.~\ref{fig:Exp3_Fusion}.
Comparison is based on the softmax loss. Two results are presented, i.e., with and without the batch normalization (BN)~\cite{ioffe2015batch} layers employed.
To enable the training of TBE-CNN by a single GPU, we only add BN layers after convolutional layers of the Inception 4 and Inception 5 modules as BN consumes video memory.
The performance of TBE-CNN is considerably higher than the trunk network.
In particular, on the more challenging handheld set, the margin is as large as 5.82\% and 3.48\% without and with BN layers, respectively.
The ROC curves of the two models on the handheld set are plotted in Fig.~\ref{fig:Exp3_TBECNNvsTrunk}.
The comparisons suggest that TBE-CNN effectively makes use of the complementary information between the holistic face image and facial components.
It is important to note that TBE-CNN is quite efficient in both time and video memory costs.
In brief, the time and memory costs of TBE-CNN are only 1.26$\times$ and 1.2$\times$ those of the trunk network, respectively.
In comparison, if there were no layer sharing, the total time and memory costs of the three networks would be 1.71$\times$ and 1.6$\times$ those of the trunk network, respectively.

\subsection{Performance Comparison on PaSC}
\begin{figure}
\centering
\includegraphics[width=1.0\linewidth]{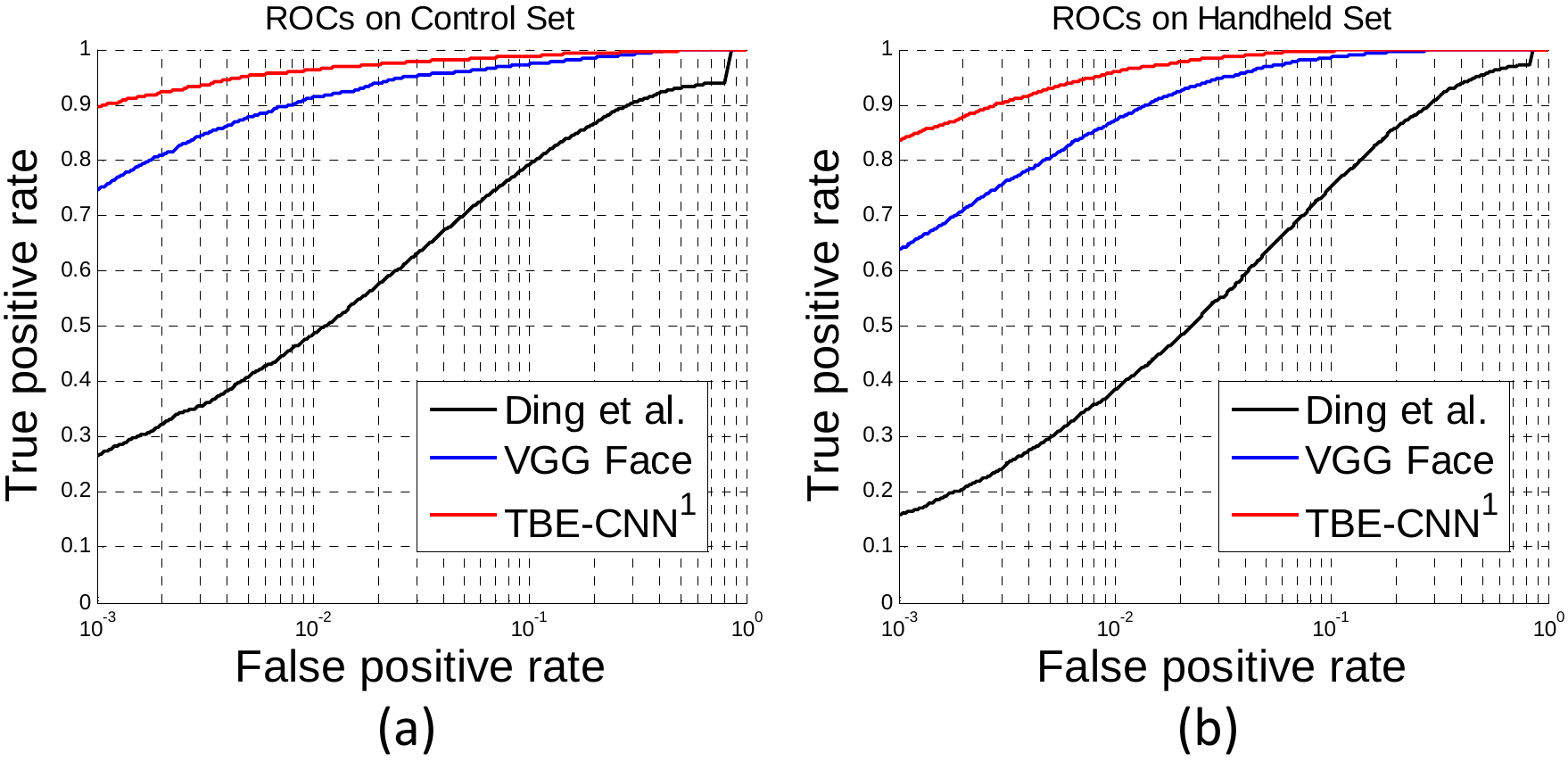}
\caption{ROC curves of TBE-CNN and state-of-the-art methods on PaSC.
The original face detection results from the database are employed for all methods.
(a) Control set; (b) handheld set.
}
\label{fig:PaSCComparison}
\end{figure}

\begin{table}[!t]
\renewcommand{\arraystretch}{1.3}
\caption{Verification Rates (\%) at 1\% FAR of Different Methods on PaSC}
\label{tab:PaSC_Performance}
\centering
\begin{tabular}{|l|c|c|c|}
\hline
                                                &\tabincell{c}{\# Training Data\\(Original)}   &\tabincell{c}{Control\\Set}     &\tabincell{c}{Handheld\\Set}\\\hline\hline
Ding \textit{et al.}~\cite{Ross2015report}      &0.01M        &48.00                  &38.00\\\hline
Huang \textit{et al.}~\cite{Ross2015report}     &0.5M         &58.00                  &59.00\\\hline
HERML~\cite{huang2015face}                      &0.5M         &46.61                  &46.23\\\hline
VGG Face~\cite{parkhi2015deep}                  &2.62M        &91.25                  &87.03\\\hline
\textbf{TBE-CNN$^1$}                            &0.49M        &\textbf{95.83}         &\textbf{94.80}\\\hline
\textbf{TBE-CNN$^1$+BN}                         &0.49M        &\textbf{96.23}         &\textbf{95.85}\\\hline
\textbf{TBE-CNN$^2$+BN}                         &0.49M        &\textbf{97.80}         &\textbf{96.12}\\\hline
\end{tabular}
\end{table}

We next present the best performance of TBE-CNN on PaSC with TS training data and MDR-TL fine-tuning.
For fair comparison with existing approaches, we present both results of TBE-CNN with and without BN layers.
Similar to the previous experiment, we only add BN layers after convolutional layers of the Inception 4 and Inception 5 modules to save video memory.
Performance comparison is illustrated in Fig.~\ref{fig:PaSCComparison} and Table~\ref{tab:PaSC_Performance}.
The performance of the VGG Face model~\cite{parkhi2015deep,simonyan2014very} is reported according to the model published by the authors.
According to the description in~\cite{parkhi2015deep}, we fine-tune the last fully connected layer of the VGG Face model with the triplet loss on CASIA-WebFace.
For VGG Face model testing, we employ the same strategy as TBE-CNN to extract video representations, as described in Section~\ref{Sec:VFR}.

In Table~\ref{tab:PaSC_Performance}, TBE-CNN$^1$ denotes the TBE-CNN performance with the original face detection results from the database.
TBE-CNN$^2$ denotes the performance based on manually cleaned face detection results.
It is clear that the proposed approach achieves the best performance on both the control and handheld sets.
In particular, the proposed approach outperforms the VGG Face model~\cite{parkhi2015deep} by 4.98\% and 8.82\% on the control and handheld sets, respectively.
It is worth noting that the VGG Face model is trained with over 2.6 million still image data and aggressive data augmentation.
In comparison, the size of our original training data is only 0.49 million.

Moreover, with the proposed techniques, we win the first place with considerable margin in the BTAS 2016 Video Person Recognition Evaluation~\cite{phillips2016report},
which is also based on the PaSC database.
With fully automatic face detection, alignment, and recognition, our four-model ensemble system achieves 98.0\% and 97.0\% verification rates at 1\% FAR on the two sets of PaSC, respectively.
The detailed comparisons on PaSC justify the effectiveness of the proposed methods for VFR.

\subsection{Performance Comparison on COX Face}
\label{sec:cox}

\begin{table*}[!t]
\renewcommand{\arraystretch}{1.3}
\caption{Rank-1 Identification Rates (\%) under the V2S/S2V Settings for Different Methods on the COX Face Database}
\label{tab:COXFace_S2V}
\centering
\begin{tabular}{|l|c|c|c||c|c|c|}
\hline
\multirow{2}{*}{}  \multirow{2}{*}{} &
\multicolumn{3}{c||}{V2S Identification Rate} &
\multicolumn{3}{c|}{S2V Identification Rate} \\
\cline{2-7}
& V1-S & V2-S & V3-S & S-V1 & S-V2 & S-V3 \\
\hline\hline
PSCL~\cite{Huang2015Benchmark}    &38.60 $\pm$ 1.39   &33.20 $\pm$ 1.77   &53.26 $\pm$ 0.80   &36.39 $\pm$ 1.61   &30.87 $\pm$ 1.77    &50.96 $\pm$ 1.44 \\\hline
LERM~\cite{huang2014learning}     &45.71 $\pm$ 2.05   &42.80 $\pm$ 1.86   &58.37 $\pm$ 3.31   &49.07 $\pm$ 1.53   &44.16 $\pm$ 0.94    &63.83 $\pm$ 1.58 \\\hline
VGG Face~\cite{parkhi2015deep}    &88.36 $\pm$ 1.02   &80.46 $\pm$ 0.76   &90.93 $\pm$ 1.02   &69.61 $\pm$ 1.46   &68.11 $\pm$ 0.91    &76.01 $\pm$ 0.71 \\\hline
\textbf{TBE-CNN}                  &\textbf{93.57 $\pm$ 0.65}   &\textbf{93.69 $\pm$ 0.51}   &\textbf{98.96 $\pm$ 0.17}  &\textbf{88.24 $\pm$ 0.40}   &\textbf{87.86 $\pm$ 0.85}    &\textbf{95.74 $\pm$ 0.67} \\\hline
\end{tabular}
\end{table*}

\begin{table*}[!t]
\renewcommand{\arraystretch}{1.3}
\caption{Rank-1 Identification Rates (\%) under the V2V Setting for Different Methods on the COX Face Database}
\label{tab:COXFace_V2V}
\centering
\begin{tabular}{|l|c|c|c||c|c|c|}
\hline
& V2-V1 & V3-V1 & V3-V2 & V1-V2 & V1-V3 & V2-V3 \\
\hline\hline
PSCL~\cite{Huang2015Benchmark}    &57.70 $\pm$ 1.40   &73.17 $\pm$ 1.44   &67.70 $\pm$ 1.70   &62.77 $\pm$ 1.02   &78.26 $\pm$ 0.97    &68.91 $\pm$ 2.28 \\\hline
LERM~\cite{huang2014learning}     &65.94 $\pm$ 1.97   &78.24 $\pm$ 1.32   &70.67 $\pm$ 1.88   &64.44 $\pm$ 1.55   &80.53 $\pm$ 1.36    &72.96 $\pm$ 1.99 \\\hline
HERML-GMM~\cite{huang2015face}    &95.10 $\pm$ -      &96.30 $\pm$ -      &94.20 $\pm$ -      &92.30 $\pm$ -      &95.40 $\pm$ -       &94.50 $\pm$ - \\\hline
VGG Face~\cite{parkhi2015deep}    &94.51 $\pm$ 0.47   &95.34 $\pm$ 0.32   &96.39 $\pm$ 0.42   &93.39 $\pm$ 0.56   &96.10 $\pm$ 0.27    &96.60 $\pm$ 0.52 \\\hline
\textbf{TBE-CNN}                  &\textbf{98.07 $\pm$ 0.32}   &\textbf{98.16 $\pm$ 0.23}   &\textbf{97.93 $\pm$ 0.20}   &\textbf{97.20 $\pm$ 0.26}   &\textbf{99.30 $\pm$ 0.16}    &\textbf{99.33 $\pm$ 0.19} \\\hline
\end{tabular}
\end{table*}

The rank-1 identification rates for TBE-CNN and state-of-the-art algorithms on the COX Face database are tabulated in Tables~\ref{tab:COXFace_S2V} and~\ref{tab:COXFace_V2V}.
In Table~\ref{tab:COXFace_S2V}, V$i$-S (S-V$i$) represents the test using the $i$-th video set as probe (gallery) and the still images as the gallery (probe).
Similarly, in Table~\ref{tab:COXFace_V2V}, V$i$-V$j$ represents the test using the $i$-th video set as probe and the $j$-th video set as the gallery.
For both TBE-CNN and VGG Face~\cite{parkhi2015deep}, we directly deploy the models in the previous experiment to evaluate their performance on COX Face.

Under all experimental settings, TBE-CNN achieves the best performance.
It is also clear that the S2V and V2S tasks are significantly more difficult than the V2V task,
suggesting a huge difference in distribution between the still image domain and video data domain.
Interestingly, TBE-CNN has overwhelming performance advantage in S2V and V2S tasks.
This may be because it is trained with both still image and simulated video data,
meaning it can learn invariant face representations from still images and video data from the same subject.
This is a great advantage for many real-world VFR tasks,
e.g., watch-lists in video surveillance applications, where the gallery is composed of high-quality ID photos and the probe includes video frames captured by surveillance cameras.
Furthermore, compared to image-set model-based VFR methods~\cite{Huang2015Benchmark,huang2015face,huang2014learning},
TBE-CNN has two key advantages: first, the extracted video representation is very compact, which means it is efficient for retrieval tasks;
and second, the representation is robust to fluctuations in the volume of frames in a video.

\subsection{Performance Comparison on YouTube Faces}
\begin{table}[!t]
\renewcommand{\arraystretch}{1.3}
\caption{Mean Verification Accuracy on the YouTube Faces Database (Restricted Protocol)}
\label{tab:YouTube_Performance}
\centering
\begin{tabular}{|l|c|c|c|}
\hline
                                      &\# Crop &\tabincell{c}{\# Training Data\\(Original)}   &Accuracy(\%) $\pm$ $S_{E}$\\\hline\hline
Deep Face$^*$~\cite{taigman2014deepface}  &1       &4.4M           &91.40 $\pm$ 1.10\\\hline
DeepID2+~\cite{sun2015deeply}     &50      &0.29M          &93.20 $\pm$ 0.20\\\hline
FaceNet~\cite{schroff2015facenet} &2       &260M           &95.12 $\pm$ 0.39\\\hline
VGG Face$^*$~\cite{parkhi2015deep}    &30      &2.62M          &91.60 $\pm$ -\\\hline
TBE-CNN$^*$                           &2       &0.49M          &93.84 $\pm$ 0.32\\\hline
\textbf{TBE-CNN}                      &2       &0.49M          &\textbf{94.96 $\pm$ 0.31}\\\hline
\end{tabular}
\end{table}

\begin{figure}
\centering
\includegraphics[width=0.95\linewidth]{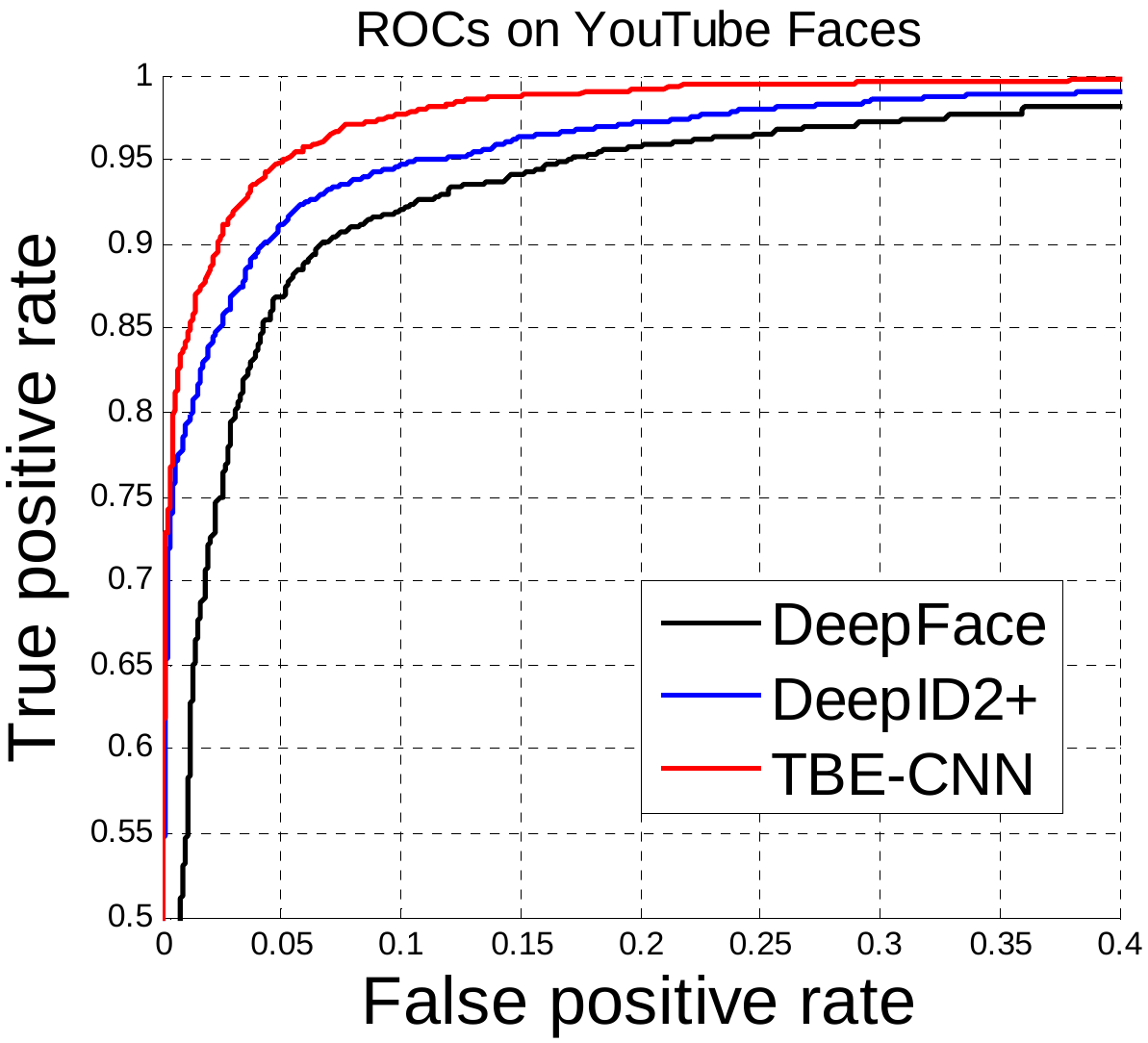}
\caption{ROC curves of TBE-CNN and state-of-the-art methods on the YouTube Faces database under the ``restricted'' protocol.
}
\label{fig:YouTubeFacesROC}
\end{figure}

Finally, we compare the face verification performance of TBE-CNN with state-of-the-art approaches on the YouTube Faces database.
As mentioned above, professional photographers rather than surveillance video cameras usually recorded videos in this database.
Therefore, they are free from image blur.
Also, since the majority of subjects in the video were in interviews, there is little pose variation, as illustrated in Fig.~\ref{fig:Database}.
Instead, the video frames are low resolution and contain serious compression artifacts.
These characteristics make experiments on YouTube Faces more similar to the traditional SIFR task rather than the real-world VFR task.
Therefore, existing SIFR models that are trained with large-scale still image databases,
e.g., the FaceNet model~\cite{schroff2015facenet}, may be expected to show advantages in this experiment.
The mean verification rates under the ``restricted'' protocol~\cite{wolf2011face} of state-of-the-art approaches are tabulated in Table~\ref{tab:YouTube_Performance}.
Corresponding ROC curves are plotted in Fig.~\ref{fig:YouTubeFacesROC}.
Under this protocol, the subject identity labels of YouTube Faces database are not allowed to be used for training.
In Table~\ref{tab:YouTube_Performance}, $^*$ denotes the models without training or fine-tuning by deep metric learning methods.

TBE-CNN outperforms previous approaches such as Deep Face~\cite{taigman2014deepface}, DeepID2+~\cite{sun2015deeply}, and VGG Face~\cite{parkhi2015deep}.
The performance of TBE-CNN is slightly lower than the FaceNet model~\cite{schroff2015facenet} that was trained with as many as 260 million face images.
Since there is no publicly available database of similar size, it will always be difficult to compete with the FaceNet model fairly.
We directly cite the performance of the VGG Face model on YouTube Faces reported in~\cite{parkhi2015deep}.
TBE-CNN$^*$ outperforms VGG Face$^*$ by as much as 2.24\%.
This result is consistent with results on the PaSC and COX Face databases.

DeepID3~\cite{sun2015deepid3} is also an important recent work that was evaluated on the LFW database~\cite{LFWTech}.
It is a huge ensemble system that includes 50 individual CNNs.
It is also worth noting that DeepID3 and TBE-CNN are designed for different face recognition tasks and therefore have their respective advantages.
The verification rate of TBE-CNN on LFW with the ``Unrestricted, Labeled Outside Data Results'' protocol~\cite{LFWTech} is $99.08\pm0.17\%$.
In comparison, the best performance of an individual CNN based on GoogLeNet in DeepID3 is about 98.68\%.
Therefore, compared with DeepID3, TBE-CNN has its own advantage.

\section{Conclusion}
\label{Sec:Conclusion}
VFR is a challenging task due to severe image blur, rich pose variations, and occlusion.
Compared to SIFR, VFR also has more demanding efficiency requirements.
Here we address these problems via a series of contributions.
First, to deal with image blur, we enrich the CNN training data by applying artificial blur to make up for the shortage of real-world video training data.
A training set composed of still images and their blurred versions encourages CNN to learn blur-robust representations.
Second, to extract pose- and occlusion-robust representations efficiently,
we propose a novel CNN architecture named TBE-CNN.
TBE-CNN efficiently extracts representations of the holistic face image and facial components by sharing the low- and middle-level layers of different CNNs.
It improves on single CNN model performance with only marginal increases in time and memory costs.
Finally, to further promote the discriminative power of the representations learnt by TBE-CNN, we propose a novel deep metric learning approach named MDR-TL,
which outperforms the widely adopted triplet loss by a considerable margin.
Extensive experiments have been conducted for S2V, V2S, and V2V tasks.
Since the proposed TBE-CNN approach effectively handles image blur, occlusion, and pose variations,
it shows clear advantages compared with state-of-the-art VFR methods on three popular video face databases.


%

%

\ifCLASSOPTIONcompsoc
  \section*{Acknowledgments}
\else
  \section*{Acknowledgment}
\fi
This work is partially supported by Australian Research Council Projects FT-130101457 and DP-140102164.




\bibliographystyle{IEEEtran}





\end{document}